\documentclass[letterpaper]{article} 
\usepackage[preprint]{aaai2027}
\usepackage[hyphens]{url}  
\usepackage{graphicx} 
\usepackage{natbib}  
\usepackage{caption} 
\usepackage{booktabs}
\usepackage{multirow}
\usepackage{amsmath}
\usepackage{amssymb}

\title{LagNet: Lag-Structured Operators for Long-Term Time Series Forecasting}

\author{
    Zhenan Yu\textsuperscript{1},
    Guangxin Jiang\textsuperscript{1},
    Jiuyun Jiang\textsuperscript{1},
    Jin Yang\textsuperscript{1}\thanks{Corresponding author.}
}
\affiliations{
    \textsuperscript{1}Harbin Institute of Technology\\
    \{25S010055@stu.hit.edu.cn, gxjiang@hit.edu.cn, jiuyunjiang@stu.hit.edu.cn, j.yang@hit.edu.cn\}
}

\newcommand{\method}{LagNet}
\newcommand{\laglayer}{LagLayer}
\newcommand{\best}[1]{\textbf{#1}}
\newcommand{\second}[1]{\underline{#1}}

\begin{document}

\maketitle

\begin{abstract}
Long-term time series forecasting maps a finite history window to an extended future horizon.
Many existing models focus primarily on the design of input representations and model structures, while paying less attention to the structural form of the history-to-future forecasting operator itself.
We examine this operator directly and find that, across model families, sensitivity matrices exhibit recurring diagonal patterns whose diagonals correspond to relative temporal lags.
Motivated by this observation, we develop a framework for direct multi-step forecasting based on lag-structured temporal operators.
Its basic building block, LagLayer, constructs a Toeplitz matrix by tying history-to-future connections with the same relative lag and incorporates horizon-wise and history-wise scaling coefficients to relax exact sharing.
This design enables LagLayer to serve as an effective replacement for temporal linear layers in existing forecasting models.
Stacking LagLayer modules with activations and multiple feature maps yields LagNet.
Experiments on long-term time series forecasting benchmarks show that our method is competitive with strong baselines, while sensitivity matrix diagnostics and limited-data experiments demonstrate that LagLayer provides an effective and sample-efficient alternative to a linear layer.
\end{abstract}

\section{Introduction}

Long-term time series forecasting (LTSF) supports decision-making in energy management, traffic planning, weather forecasting, finance, and industrial monitoring.
Compared with short-horizon forecasting, LTSF maps a finite historical window to an extended forecast horizon, placing greater demands on temporal modeling, generalization, and computational efficiency.
Recent methods largely differ in how they represent and mix historical observations.
Transformer variants target long-sequence modeling~\cite{zhou2021informer,wu2021autoformer,zhou2022fedformer}, while PatchTST~\cite{nie2023patchtst} and iTransformer~\cite{liu2024itransformer} reorganize the input into temporal patches and variate tokens, respectively.
Linear and MLP models provide simpler alternatives for temporal mixing~\cite{zeng2023dlinear,chen2023tsmixer,wang2024timemixer}.
Regardless of architecture, each forecaster induces a mapping from the historical window to the future horizon.
A complementary question is \emph{whether the history-to-future forecasting operator contains reusable structure and, if so, how that structure should be parameterized}.

\begin{figure}[t]
\centering
\includegraphics[width=0.95\columnwidth,draft=false]{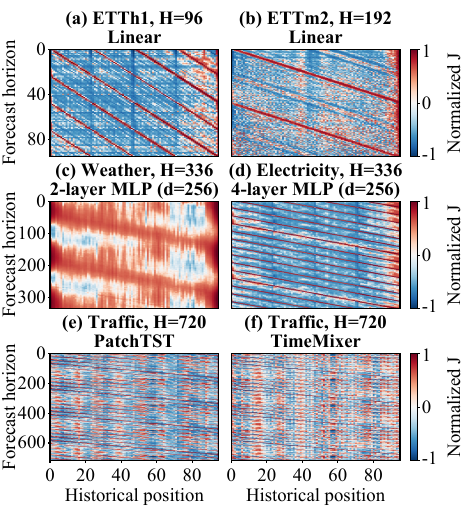}
\caption{History-to-future sensitivity matrices for six forecasting settings. Each panel averages input--output Jacobians over test windows and channels.}
\label{fig:forecasting-matrix-evidence}
\end{figure}

To compare linear and nonlinear forecasters in a common form, we examine their input-to-output Jacobians, a standard gradient-based sensitivity measure~\cite{simonyan2014saliency,ismail2020benchmarking}.
For a forecaster $f$, the Jacobian entry $J_{h,\ell}=\partial f_h/\partial x_\ell$ measures the local influence of historical position $\ell$ on forecast step $h$.
Averaging these sensitivities over test windows and channels produces an $H\times L$ matrix of influential history-to-future connections, where $H$ and $L$ denote the forecast horizon and history length, respectively.
For a linear model, $f(\boldsymbol{x})=\boldsymbol{W}\boldsymbol{x}+\boldsymbol{b}$ and $\boldsymbol{J}=\boldsymbol{W}$, so this diagnostic recovers the learned forecasting matrix $\boldsymbol{W}\in\mathbb{R}^{H\times L}$, whose $H\times L$ coefficients are fitted independently.
We refer to this one-layer model as Linear.

A connection between historical position $\ell$ and forecast step $h$ spans the relative lag $d=L+h-\ell$.
Connections with the same lag therefore lie on the same matrix diagonal.
Figure~\ref{fig:forecasting-matrix-evidence}(a) and (b) show pronounced diagonal bands in two Linear settings, with strong sensitivities recurring at approximately periodic lags.
As shown in Figure~\ref{fig:forecasting-matrix-evidence} (c)--(f), similar lag-aligned patterns appear in the MLP, PatchTST, and TimeMixer settings, although their strength and axis-aligned variation differ across panels.
Their recurrence across architectures suggests that relative lag systematically influences which history-to-future connections receive high importance.
This observation motivates our central hypothesis: connections with the same relative lag often carry similar predictive meaning, consistent with relative temporal invariance under valid time shifts.

We encode this dominant organization in a lag-structured matrix by sharing one base coefficient among connections with the same relative lag.
This yields a rectangular Toeplitz matrix~\cite{gray2006toeplitz} with $L+H-1$ lag weights instead of $H\times L$ independently learned coefficients.
The empirical sensitivities, however, are not exactly Toeplitz: several panels also exhibit horizontal variation across forecast steps and vertical concentration at selected historical positions.
These patterns suggest that relative lag provides the structural backbone, while forecast step and historical position further modulate connection strength.
Accordingly, \laglayer{} modulates the lag-structured matrix with horizon-wise and history-wise scale coefficients.
\method{} stacks these operators with pointwise nonlinearities and multiple hidden feature maps, adding capacity while retaining lag structure in every temporal mapping.

Our contributions are summarized as follows:
\begin{itemize}
    \item We identify recurring lag-aligned structure in history-to-future sensitivity matrices across diverse forecasting architectures without explicit lag sharing. This evidence motivates a relative-lag sharing prior for direct multi-step forecasting.
    \item We develop a model family around this prior. \laglayer{} combines a lag-structured matrix with horizon-wise and history-wise adaptation, while \method{} extends the operator through nonlinear depth and multiple hidden feature maps.
    \item We demonstrate competitive performance against strong baselines on eight standard LTSF benchmarks. Ablations isolate the effects of adaptive scaling and nonlinear capacity, while matrix diagnostics and limited-data experiments provide complementary evidence for relative-lag sharing. Matched replacement experiments with existing architectures further examine operator transferability.
\end{itemize}

\section{Related Work}

\subsection{Representation and Architecture Design for Long-Term Forecasting}
Long-term forecasting models have primarily advanced through new representations and architectures.
Transformer-based methods improve long-range modeling through sparse attention, decomposition, and frequency-domain processing~\cite{zhou2021informer,wu2021autoformer,zhou2022fedformer}, while PatchTST~\cite{nie2023patchtst} and iTransformer~\cite{liu2024itransformer} reorganize temporal and variate dimensions through patch and variate tokens.
TimesNet~\cite{wu2023timesnet} converts discovered periods into two-dimensional representations, and convolutional models capture temporal patterns with hierarchical or large-receptive-field designs~\cite{liu2022scinet,luo2024moderntcn}.

Direct forecasters provide a simpler alternative.
N-BEATS and NHITS construct long-horizon predictions through deep basis expansion and hierarchical interpolation~\cite{oreshkin2020nbeats,challu2023nhits}, while DLinear~\cite{zeng2023dlinear} demonstrates the competitiveness of a single linear temporal mapping.
TiDE~\cite{das2023tide}, TSMixer~\cite{chen2023tsmixer,ekambaram2023tsmixer}, and TimeMixer~\cite{wang2024timemixer} develop MLP-based encoders or mixing architectures.
Recent lightweight models further reduce cost through simplified mixing, sparse or frequency-domain parameterization, and factorized mappings~\cite{chen2025simpletm,lin2024sparsetsf,xu2024fits,ma2026mixlinear}.

These works mainly refine input representations, feature mixing, or compact parameterizations, while the structure of the direct history-to-future mapping remains less explored.
We encode the recurring relative-lag structure observed in this operator's sensitivity matrices through \laglayer{} and its nonlinear extension, \method{}, reducing the parameters required by each temporal mapping.

\subsection{Structured Operators and Toeplitz Constraints}
Structured operators have long supported efficient sequence modeling.
Temporal convolution reuses weights across relative local positions~\cite{bai2018tcn}, while structured state-space models admit convolutional representations for long-range dependencies~\cite{gu2022s4}.
Toeplitz matrices provide the corresponding matrix form for linear time-invariant filtering~\cite{gray2006toeplitz}.
Within this broader family, our setting differs from local temporal convolution: rather than sliding a kernel within a sequence, our lag-structured matrix directly maps a history window to a distinct forecast horizon.

Most closely related to our work, the Toeplitz Neural Network (TNN) employs relative-position-encoded square Toeplitz operators for within-sequence token mixing, focusing on computational efficiency and length extrapolation in general sequence modeling~\cite{qin2023tnn}.
We adopt a related sharing principle in a rectangular Toeplitz matrix, but derive it from lag-aligned patterns observed in long-term forecasters and use it to directly constrain the history-to-future mapping.
TNN therefore studies positional mixing within a sequence, whereas we study the forecasting operator between history and future.

\begin{figure}[t]
\centering
\includegraphics[width=0.90\columnwidth,draft=false]{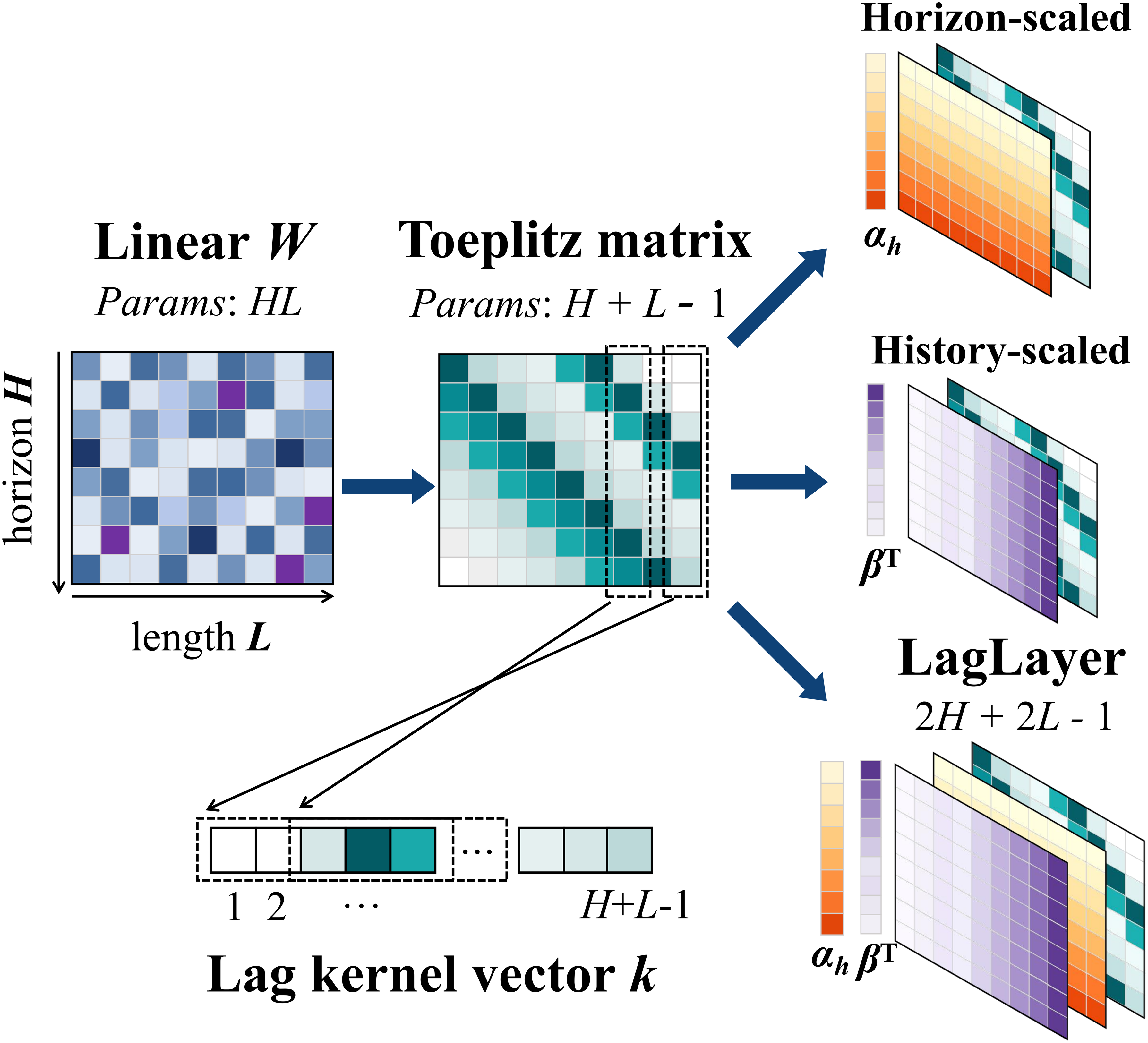}
\caption{Linear, lag-structured, and adaptively scaled forecasting matrices.}
\label{fig:matrix-comparison}
\end{figure}

\section{Method}

Consider a multivariate time series with $C$ channels.
For each channel $c\in\{1,\ldots,C\}$, the historical sequence and
future target are
\begin{equation*}
\begin{aligned}
    \boldsymbol{x}^{(c)}_{t-L+1:t}
    &=
    [x^{(c)}_{t-L+1},\ldots,x^{(c)}_t]^\top
    \in\mathbb{R}^{L}, \\
    \boldsymbol{x}^{(c)}_{t+1:t+H}
    &=
    [x^{(c)}_{t+1},\ldots,x^{(c)}_{t+H}]^\top
    \in\mathbb{R}^{H},
\end{aligned}
\end{equation*}
where $x_t^{(c)}\in\mathbb{R}$ is the observation of channel $c$ at
time $t$, $L$ is the history length, and $H$ is the forecast horizon.

\subsection{Lag-Structured Forecasting Operator}
We derive the temporal operator for an arbitrary channel $c$ and omit the superscript $(c)$ for readability.
Parameter sharing across channels is specified separately for each model configuration.

A dense forecasting operator uses a matrix $\boldsymbol{W}\in\mathbb{R}^{H\times L}$, where $w_{h,\ell}=[\boldsymbol{W}]_{h,\ell}$ is its $(h,\ell)$-th element.
Given the historical sequence, the dense prediction is
\begin{equation*}
    \widehat{\boldsymbol{x}}_{t+1:t+H}
    =
    \boldsymbol{W}\boldsymbol{x}_{t-L+1:t}
    +
    \boldsymbol{b}.
\end{equation*}
Here, $\boldsymbol{b}\in\mathbb{R}^{H}$, and all $HL$ elements of $\boldsymbol{W}$ are learned independently.

The element $w_{h,\ell}$ connects the historical value $x_{t-L+\ell}$ to the prediction $\hat{x}_{t+h}$, with relative lag $(t+h)-(t-L+\ell)=L+h-\ell$.
A dense matrix learns these connections independently, even when they
have the same relative lag.
The recurring diagonal patterns in
Figure~\ref{fig:forecasting-matrix-evidence} suggest that connections
with the same lag can share a common base coefficient.
We can therefore construct a lag-structured matrix.
Let $\boldsymbol{k}=[k_1,\ldots,k_{L+H-1}]^\top\in\mathbb{R}^{L+H-1}$ contain one coefficient for each possible lag, where $k_1$ corresponds to the shortest lag and $k_{L+H-1}$ to the longest.
We define
\begin{equation}
    w_{h,\ell}
    =
    k_{L+h-\ell}.
    \label{eq:lag-sharing}
\end{equation}
Let $\mathcal{T}_{H,L}$ denote this construction.
The resulting lag-structured matrix is
\begin{equation}
\begin{aligned}
\boldsymbol{W}_{\mathrm{lag}}
&=
\mathcal{T}_{H,L}(\boldsymbol{k}) \\
&=
\begin{bmatrix}
k_L & k_{L-1} & \cdots & k_1 \\
k_{L+1} & k_L & \cdots & k_2 \\
\vdots & \vdots & \ddots & \vdots \\
k_{L+H-1} & k_{L+H-2} & \cdots & k_H
\end{bmatrix}.
\end{aligned}
\label{eq:lag-matrix}
\end{equation}
Because its entries depend only on $h-\ell$,
$\boldsymbol{W}_{\mathrm{lag}}$ is a rectangular Toeplitz matrix.
The corresponding unscaled operator applies this matrix to the historical sequence:
\begin{equation*}
    \widehat{\boldsymbol{x}}_{t+1:t+H}
    =
    \boldsymbol{W}_{\mathrm{lag}}
    \boldsymbol{x}_{t-L+1:t}
    +
    \boldsymbol{b}.
\end{equation*}

This parameterization uses $L+H-1$ lag coefficients instead of the
$HL$ independently learned elements of $\boldsymbol{W}$.

\subsection{\laglayer{} with Adaptive Scaling}

\begin{figure}[t]
\centering
\includegraphics[width=\columnwidth,draft=false]{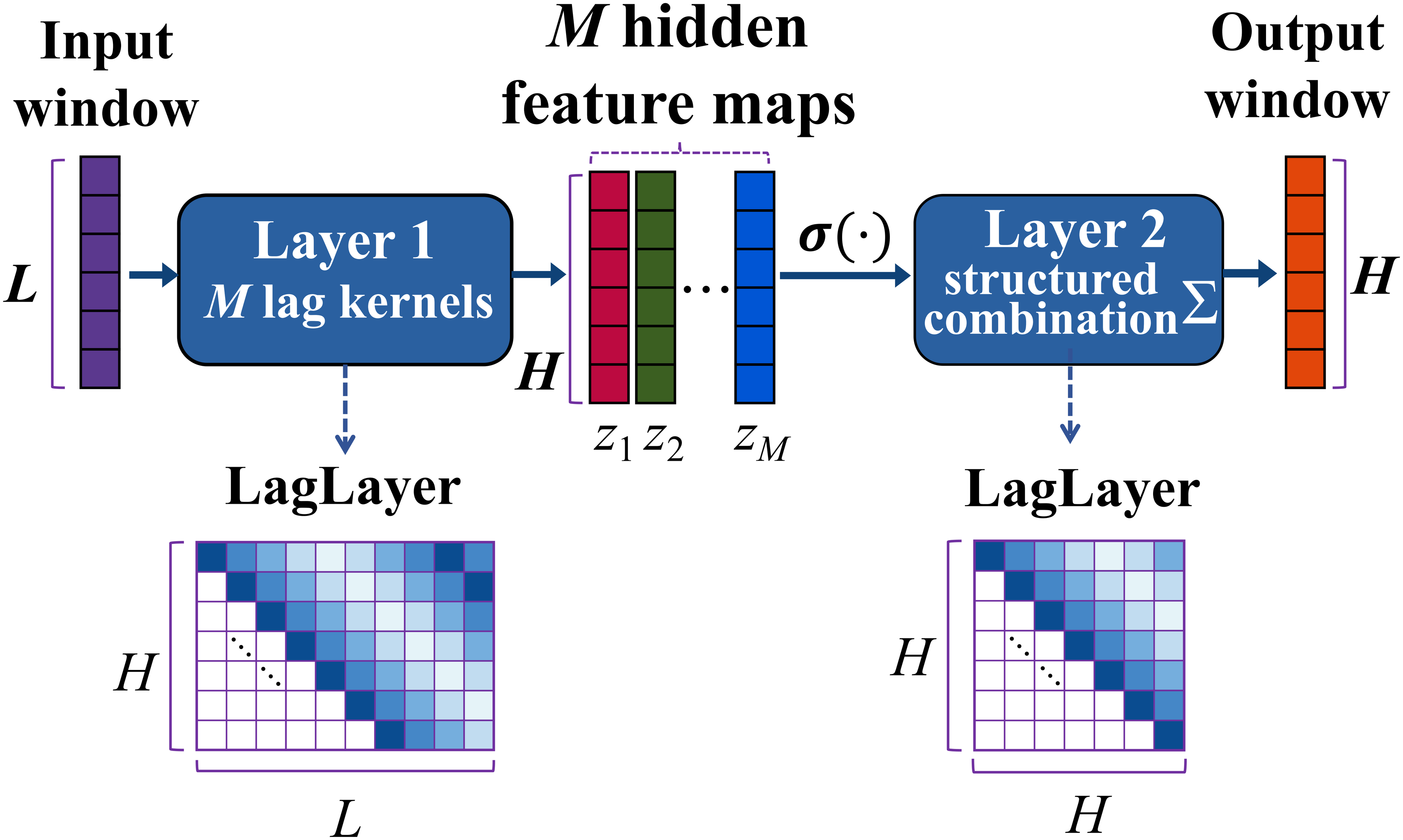}
\caption{Two-layer \method{} architecture. Deeper models repeat a \laglayer{} followed by a pointwise activation, and hidden feature maps allow the network to represent multiple lag-response patterns.}
\label{fig:lagnet-architecture}
\end{figure}

Exact lag sharing assigns one coefficient to all connections on the
same diagonal.
The sensitivity matrices in
Figure~\ref{fig:forecasting-matrix-evidence} retain a diagonal
organization but also vary across forecast steps and historical
positions.
Several panels also contain vertically concentrated regions of high sensitivity,
including pronounced responses near the newest observations.
This variation indicates that relative lag alone does not fully
describe the forecasting matrix.

To capture this variation without discarding the relative lag
structure, \laglayer{} introduces a horizon-wise scale vector
$\boldsymbol{\alpha}\in\mathbb{R}^{H}$ and a
history-wise scale vector
$\boldsymbol{\beta}\in\mathbb{R}^{L}$.
For a connection from historical position $\ell$ to forecast step
$h$, the scaled coefficient is
\begin{equation*}
\begin{aligned}
    w_{h,\ell}
    &= \alpha_h k_{L+h-\ell}
    && \text{(horizon-wise)}, \\
    w_{h,\ell}
    &= \beta_\ell k_{L+h-\ell}
    && \text{(history-wise)}, \\
    w_{h,\ell}
    &= \alpha_h\beta_\ell k_{L+h-\ell}
    && \text{(both)}.
\end{aligned}
\end{equation*}
Here, $\alpha_h$ modulates the corresponding forecast row, while
$\beta_\ell$ modulates the corresponding historical column.
The lag coefficient $k_{L+h-\ell}$ retains the shared temporal
backbone.

Let $\boldsymbol{D}_{\alpha}=\operatorname{diag}(\boldsymbol{\alpha})\in\mathbb{R}^{H\times H}$ and $\boldsymbol{D}_{\beta}=\operatorname{diag}(\boldsymbol{\beta})\in\mathbb{R}^{L\times L}$ denote the corresponding diagonal matrices.
When both scale vectors are enabled, the effective \laglayer{} matrix is
\begin{equation}
\begin{aligned}
\boldsymbol{W}_{\mathrm{layer}}
&=
\boldsymbol{D}_{\alpha}
\mathcal{T}_{H,L}(\boldsymbol{k})
\boldsymbol{D}_{\beta} \\
&=
\boldsymbol{D}_{\alpha}
\begin{bmatrix}
\beta_1 k_L
& \beta_2 k_{L-1}
& \cdots
& \beta_L k_1 \\
\beta_1 k_{L+1}
& \beta_2 k_L
& \cdots
& \beta_L k_2 \\
\vdots & \vdots & \ddots & \vdots \\
\beta_1 k_{L+H-1}
& \beta_2 k_{L+H-2}
& \cdots
& \beta_L k_H
\end{bmatrix}.
\end{aligned}
    \label{eq:laglayer-matrix}
\end{equation}
Unlike its Toeplitz backbone, $\boldsymbol{W}_{\mathrm{layer}}$ is
generally not Toeplitz because entries on the same diagonal can differ
through $\alpha_h\beta_\ell$.
The shared structure therefore applies to the base lag coefficients,
rather than requiring identical effective weights along each diagonal.

The corresponding prediction is
\begin{equation}
\begin{aligned}
    \widehat{\boldsymbol{x}}_{t+1:t+H}
    &=
    \boldsymbol{W}_{\mathrm{layer}}\boldsymbol{x}_{t-L+1:t}
    +
    \boldsymbol{b} \\
    &=
    \boldsymbol{D}_{\alpha}
    \mathcal{T}_{H,L}(\boldsymbol{k})
    \boldsymbol{D}_{\beta}
    \boldsymbol{x}_{t-L+1:t}
    +
    \boldsymbol{b}.
\end{aligned}
    \label{eq:laglayer-prediction}
\end{equation}
The scale coefficients are initialized to one, so \laglayer{} initially recovers its unscaled lag-structured backbone. They subsequently provide controlled flexibility along the forecast-horizon and historical-position axes without removing that backbone.
Throughout the main experiments, \laglayer{} denotes the complete single-layer operator with both scale vectors; configurations with either or both scales removed are used only for ablation.
Figure~\ref{fig:matrix-comparison} summarizes the difference between the dense, lag-structured, and adaptively scaled parameterizations.

\subsection{Deep Lag-Structured Networks}
A single \laglayer{} is linear.
\method{} introduces nonlinear capacity by stacking $R\geq2$
\laglayer{} modules with pointwise activations, as shown in
Figure~\ref{fig:lagnet-architecture}.

We use a unified notation for all layers.
Let $T_r$ and $M_r$ denote the temporal length and number of feature
maps at layer $r$, respectively:
\begin{equation*}
\begin{aligned}
    T_0 &= L, &
    T_r &= H, && 1\leq r\leq R, \\
    M_0 &= 1, &
    M_R &= 1, &
    M_r &= M, && 1\leq r<R.
\end{aligned}
\end{equation*}
The input and output are treated as single feature maps.
Each hidden layer contains $M$ latent maps whose temporal axis has
length $H$; these latent maps are distinct from the $C$ observed
channels.
This choice aligns the hidden temporal positions with the forecast
steps; the hidden maps are latent representations rather than
individually supervised forecasts.
The extension to a general hidden length $H'$ is given in the
supplementary material.

For layer $r$, each input-output map pair has its own lag kernel and
scale vectors:
\begin{equation}
    \boldsymbol{W}^{(r)}_{m,n}
    =
    \boldsymbol{D}^{(r)}_{\alpha,m,n}
    \mathcal{T}_{T_r,T_{r-1}}
    \!\left(\boldsymbol{k}^{(r)}_{m,n}\right)
    \boldsymbol{D}^{(r)}_{\beta,m,n}
    \in
    \mathbb{R}^{T_r\times T_{r-1}}.
    \label{eq:lagnet-matrices}
\end{equation}
Here,
$m=1,\ldots,M_r$ and $n=1,\ldots,M_{r-1}$.
The lag kernel has length $T_r+T_{r-1}-1$.
The corresponding row- and column-wise diagonal scale matrices satisfy
$\boldsymbol{D}^{(r)}_{\alpha,m,n}\in\mathbb{R}^{T_r\times T_r}$
and
$\boldsymbol{D}^{(r)}_{\beta,m,n}\in\mathbb{R}^{T_{r-1}\times T_{r-1}}$,
respectively.

Let
$\boldsymbol{z}^{(r)}_m\in\mathbb{R}^{T_r}$
denote map $m$ at layer $r$, and let
$\boldsymbol{b}^{(r)}_m\in\mathbb{R}^{T_r}$
denote its bias.
The network is computed as
\begin{equation}
\begin{aligned}
    \boldsymbol{z}^{(r)}_m
    &=
    \sigma_r\!\left(
    \sum_{n=1}^{M_{r-1}}
    \boldsymbol{W}^{(r)}_{m,n}
    \boldsymbol{z}^{(r-1)}_n
    +
    \boldsymbol{b}^{(r)}_m
    \right), \\
    &\qquad
    1\leq r\leq R,\quad
    1\leq m\leq M_r,
\end{aligned}
    \label{eq:lagnet-recurrence}
\end{equation}
with
$\boldsymbol{z}^{(0)}_1=\boldsymbol{x}_{t-L+1:t}$
and
$\boldsymbol{z}^{(R)}_1=
\widehat{\boldsymbol{x}}_{t+1:t+H}$.
The hidden-layer functions $\sigma_r$, $r<R$, are pointwise
activations, while $\sigma_R$ is the identity map.

The first layer maps the history to $M$ future-aligned latent maps.
Each intermediate layer mixes these maps through $M^2$ pair-specific
\laglayer{} operators, and the final layer returns one forecast.
Depth composes structured transformations with nonlinearities, while
the feature-map dimension allows the network to represent multiple
lag-response patterns without changing the hidden temporal axis.

\subsection{Parameter Complexity}
Table~\ref{tab:complexity} summarizes the parameter complexity of single-channel forecasting operators, excluding biases.
Compared with the $HL$ independent parameters of a dense operator,
\laglayer{} uses $L+H-1$ lag coefficients and $H+L$ horizon- and
history-wise scales, totaling $2L+2H-1$ parameters without recovering
$HL$ independent weights.

\begin{table}[t]
\centering
{
\setlength{\tabcolsep}{1mm}
\begin{tabular}{lcc}
\toprule
Operator & Element form & Parameters \\
\midrule
Linear & $w_{h,\ell}$ & $HL$ \\
Lag-structured matrix & $k_{L+h-\ell}$ & $L+H-1$ \\
Horizon scale only & $\alpha_h k_{L+h-\ell}$ & $L+2H-1$ \\
History scale only & $\beta_\ell k_{L+h-\ell}$ & $2L+H-1$ \\
\laglayer{} & $\alpha_h\beta_\ell k_{L+h-\ell}$ & $2L+2H-1$ \\
\bottomrule
\end{tabular}
}
\caption{Parameter complexity of forecasting operators, with bias terms omitted for clarity.}
\label{tab:complexity}
\end{table}

\begin{table*}[t]
\centering
{\small
\setlength{\tabcolsep}{1mm}
\begin{tabular*}{\textwidth}{@{\extracolsep{\fill}}l*{18}{c}@{}}
\toprule
& \multicolumn{10}{c}{Baselines} & \multicolumn{8}{c}{Proposed} \\
\cmidrule(lr){2-11}\cmidrule(lr){12-19}
Dataset
& \multicolumn{2}{c}{\shortstack{DLinear\\(2023)}}
& \multicolumn{2}{c}{\shortstack{PatchTST\\(2023)}}
& \multicolumn{2}{c}{\shortstack{iTransformer\\(2024)}}
& \multicolumn{2}{c}{\shortstack{TimeMixer\\(2024)}}
& \multicolumn{2}{c}{\shortstack{TimeKAN\\(2025)}}
& \multicolumn{2}{c}{\shortstack{\laglayer{}\\(ours)}}
& \multicolumn{2}{c}{\shortstack{\method{}\\(ours)}}
& \multicolumn{2}{c}{\shortstack{Lag-Time\\Mixer (ours)}}
& \multicolumn{2}{c}{\shortstack{Lag-Time\\KAN (ours)}} \\
\cmidrule(lr){2-3}\cmidrule(lr){4-5}\cmidrule(lr){6-7}\cmidrule(lr){8-9}
\cmidrule(lr){10-11}\cmidrule(lr){12-13}\cmidrule(lr){14-15}\cmidrule(lr){16-17}\cmidrule(lr){18-19}
& MSE & MAE & MSE & MAE & MSE & MAE & MSE & MAE
& MSE & MAE & MSE & MAE & MSE & MAE & MSE & MAE & MSE & MAE \\
\midrule
ETTh1
& 0.463 & 0.458 & 0.516 & 0.484 & 0.454 & 0.448
& 0.464 & 0.448 & \second{0.422} & \second{0.429} & 0.449 & 0.435 & 0.461 & 0.443 & 0.459 & 0.444 & \best{0.421} & \best{0.428} \\
ETTh2
& 0.512 & 0.491 & 0.391 & 0.412 & 0.383 & 0.407
& \second{0.376} & 0.404 & \second{0.376} & \second{0.401} & 0.378 & \best{0.399} & 0.384 & 0.408 & \best{0.373} & \second{0.401} & 0.386 & 0.409 \\
ETTm1
& 0.405 & 0.409 & 0.406 & 0.408 & 0.407 & 0.410
& 0.384 & \second{0.398} & \best{0.381} & 0.399 & 0.410 & 0.404 & 0.388 & 0.402 & \second{0.383} & \best{0.397} & 0.387 & 0.402 \\
ETTm2
& 0.321 & 0.371 & 0.290 & 0.334 & 0.288 & 0.332
& \second{0.279} & \second{0.325} & 0.284 & 0.329 & 0.285 & 0.326 & 0.281 & \second{0.325} & \best{0.277} & \best{0.324} & 0.292 & 0.339 \\
Weather
& 0.267 & 0.317 & 0.265 & 0.286 & 0.258 & 0.278
& 0.246 & 0.276 & \best{0.244} & \second{0.274} & 0.246 & \best{0.272} & 0.249 & 0.278 & \second{0.245} & 0.276 & 0.248 & 0.280 \\
Solar
& 0.326 & 0.397 & 0.287 & 0.333 & 0.233 & \best{0.262}
& \best{0.230} & 0.298 & 0.282 & 0.315 & 0.368 & 0.359 & 0.255 & \second{0.282} & \second{0.233} & 0.303 & 0.311 & 0.339 \\
Electricity
& 0.210 & 0.296 & 0.216 & 0.318 & \best{0.178} & \best{0.270}
& \second{0.183} & \second{0.273} & 0.199 & 0.289 & 0.219 & 0.295 & 0.191 & 0.278 & 0.186 & 0.278 & 0.203 & 0.292 \\
Traffic
& 0.625 & 0.384 & 0.529 & 0.341 & \best{0.428} & \best{0.282}
& 0.491 & 0.303 & 0.580 & 0.372 & 0.632 & 0.378 & \second{0.458} & \second{0.293} & 0.499 & 0.306 & 0.595 & 0.374 \\
\bottomrule
\end{tabular*}
}
\caption{Four-horizon average forecasting results for $L=96$. All results are averaged over five random seeds. The best and second-best values are shown in bold and underlined, respectively; complete horizon-wise results and seed statistics are provided in the supplementary material.}
\label{tab:main-results}
\end{table*}

\begin{table*}[t]
\centering
{\small
\setlength{\tabcolsep}{1.5mm}
\begin{tabular}{lcccccc}
\toprule
& \multicolumn{2}{c}{DLinear $\rightarrow$ DLagLayer}
& \multicolumn{2}{c}{TimeMixer $\rightarrow$ Lag-TimeMixer}
& \multicolumn{2}{c}{TimeKAN $\rightarrow$ Lag-TimeKAN} \\
\cmidrule(lr){2-3}\cmidrule(lr){4-5}\cmidrule(lr){6-7}
Dataset & $\Delta$ MSE (\%) & Params
& $\Delta$ MSE (\%) & Params
& $\Delta$ MSE (\%) & Params \\
\midrule
ETT (Avg.) & $-1.3$ & 65.2K $\to$ 2.4K & $-0.6$ & 120.1K $\to$ 62.7K & $+1.7$ & 41.5K $\to$ 10.1K \\
Weather & $+2.2$ & 65.2K $\to$ 2.4K & $-0.1$ & 119.7K $\to$ 62.3K & $+1.8$ & 41.2K $\to$ 9.8K \\
Solar & $+1.5$ & 65.2K $\to$ 2.4K & $+1.0$ & 2.37M $\to$ 2.25M & $+10.4$ & 45.1K $\to$ 13.8K \\
Electricity & $+1.8$ & 65.2K $\to$ 2.4K & $+1.2$ & 122.1K $\to$ 64.7K & $+2.1$ & 46.6K $\to$ 15.2K \\
Traffic & $+0.9$ & 65.2K $\to$ 2.4K & $+1.8$ & 132.8K $\to$ 75.4K & $+2.7$ & 50.9K $\to$ 19.6K \\
\midrule
Average & $+0.1$ & $-96.3\%$ & $+0.2$ & $-41.8\%$ & $+3.0$ & $-72.8\%$ \\
\bottomrule
\end{tabular}
}
\caption{Paired operator replacement over four horizons. Negative values indicate reductions. Trainable parameter counts are reported before and after replacement. ETT (Avg.) averages the four ETT datasets.}
\label{tab:param-summary}
\end{table*}

\begin{figure*}[t]
\centering
\includegraphics[width=0.98\textwidth,draft=false]{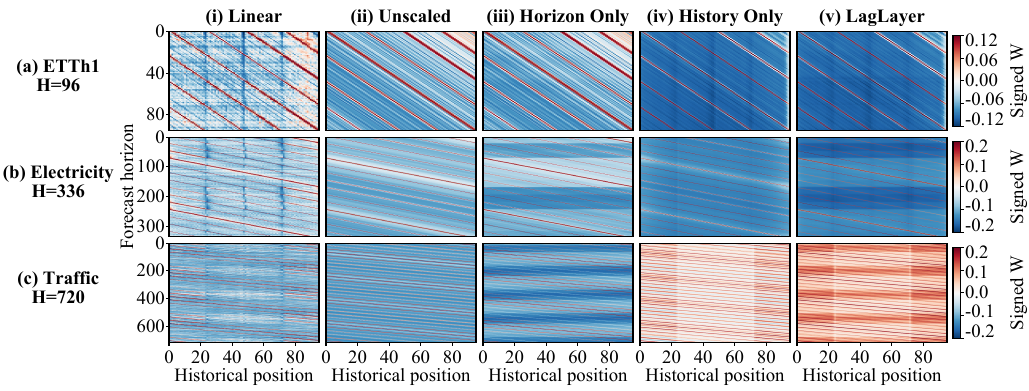}
\caption{Signed effective forecasting matrices for Linear and lag-structured variants. Heatmaps within each row share a symmetric color scale whose magnitude limit is the 99.5th percentile of absolute weight values.}
\label{fig:matrix-visualization}
\end{figure*}

The parameter reduction also applies to hidden-to-hidden transformations.
A shape-matched dense MLP flattens an $H\times M$ representation to
width $HM$ and requires $\mathcal{O}(M^2H^2)$ parameters per hidden
layer.
The corresponding \method{} layer contains $M^2$ pair-specific
\laglayer{} operators, each with $\mathcal{O}(H)$ kernel and scale
parameters.
Its total parameter count is therefore $\mathcal{O}(M^2H)$.
Lag structuring reduces the parameter dependence on the temporal
length from quadratic to linear while retaining all-to-all mixing
between feature maps.

\section{Experiments}

\subsection{Experimental Setup}
\paragraph{Datasets.}
We evaluate on eight standard LTSF benchmarks: ETTh1, ETTh2, ETTm1, ETTm2,
Electricity, Solar-Energy, Traffic, and Weather. We follow their public
chronological splits and established
preprocessing~\cite{zeng2023dlinear,wu2023timesnet,nie2023patchtst,liu2024itransformer,wang2024timemixer}.
Dataset statistics and exact split sizes are provided in the supplementary
material.

\paragraph{Baselines and Experimental Settings.}
We compare with DLinear~\cite{zeng2023dlinear}, PatchTST~\cite{nie2023patchtst},
iTransformer~\cite{liu2024itransformer}, TimeMixer~\cite{wang2024timemixer},
and TimeKAN~\cite{huang2025timekan}.
They cover linear, Transformer, multiscale MLP, and frequency-decomposition
architectures.
All experiments use $L=96$ and $H\in\{96,192,336,720\}$ under a unified
setting, with results averaged over five random seeds.
We report MSE as the primary metric and MAE as a complementary metric; Params
denotes the total number of trainable parameters.

\paragraph{Proposed Models and Implementation.}
We evaluate \laglayer{} and \method{}, together with DLinear,
TimeMixer, and TimeKAN transfer variants.
Each variant replaces only the linear temporal forecasting mapping with
\laglayer{} while retaining all other components and training settings, without
specific tuning.
We use PyTorch, Adam, and MSE loss; adaptive scales are
initialized to one, and testing uses the checkpoint with the lowest validation
MSE.
The supplementary material provides the complete settings and results.

\subsection{Main Results}
Table~\ref{tab:main-results} compares the four proposed models with five baselines across eight datasets.
At least one proposed model ranks first in MSE or MAE on five datasets and
among the top two on seven.
\laglayer{} achieves the lowest MAE on ETTh2 and Weather, while the transferred
models take first place in at least one metric on every ETT dataset.
Electricity is the only dataset on which no proposed model ranks among the
top two.

The largest gains from nonlinear composition occur on Solar, Electricity, and
Traffic.
Relative to \laglayer{}, \method{} reduces average MSE by 30.8\%, 13.0\%, and
27.6\% on these datasets, respectively.
It also ranks second in both MSE and MAE on Traffic.
Thus, added depth can be particularly beneficial for high-dimensional
multivariate datasets.

Table~\ref{tab:param-summary} isolates the effect of replacing each backbone's
linear temporal forecasting mapping with \laglayer{} while leaving the other components
unchanged.
This replacement reduces parameter counts by 96.3\% for DLinear, 41.8\% for
TimeMixer, and 72.8\% for TimeKAN.
The corresponding increases in average MSE are 0.1\%, 0.2\%, and 3.0\%.
The largest accuracy cost occurs for TimeKAN on Solar, where MSE increases by
10.4\%.
The replacement therefore preserves average accuracy closely for DLinear and
TimeMixer, but less consistently for TimeKAN.
Results across these three models, which employ distinct architectures and
mechanisms, further indicate that relative-lag sharing is a transferable
operator prior rather than one tied to a particular forecasting backbone,
substantially reducing parameter counts while retaining broadly comparable
aggregate forecasting accuracy.

\subsection{Sample Efficiency Analysis}
Relative-lag sharing provides an effective structural prior that reduces
the number of forecasting coefficients that must be estimated independently
and may therefore improve sample efficiency.
We evaluate Linear, \laglayer{}, and \method{} using 1\%, 5\%, 20\%,
50\%, or 100\% of the original training data, while keeping the
validation and test sets and all other training settings unchanged.
Table~\ref{tab:sample-efficiency} reports the average MSE over the four
prediction lengths.
The main table aggregates the four ETT datasets, with complete results
provided in the supplementary material.

\begin{table}[t]
\centering
{\small
\setlength{\tabcolsep}{1mm}
\begin{tabular}{@{}llcccccc@{}}
\toprule
Dataset & Model & 1\% & 5\% & 20\% & 50\% & 100\% & Avg. \\
\midrule
\multirow{3}{*}{ETT (Avg.)} & Linear & 0.432 & 0.399 & \best{0.385} & \best{0.383} & \best{0.381} & 0.396 \\
 & LagLayer & \best{0.401} & 0.392 & 0.386 & 0.386 & 0.385 & 0.390 \\
 & LagNet & \best{0.401} & \best{0.386} & 0.389 & 0.385 & 0.385 & \best{0.389} \\
\cmidrule{1-8}
\multirow{3}{*}{Weather} & Linear & 0.287 & \best{0.275} & 0.273 & 0.272 & 0.272 & 0.276 \\
 & LagLayer & 0.285 & 0.277 & 0.274 & 0.273 & 0.272 & 0.276 \\
 & LagNet & \best{0.281} & 0.276 & \best{0.270} & \best{0.265} & \best{0.265} & \best{0.271} \\
\cmidrule{1-8}
\multirow{3}{*}{Solar} & Linear & 0.431 & 0.382 & 0.363 & 0.358 & 0.356 & 0.378 \\
 & LagLayer & 0.540 & 0.444 & 0.389 & 0.373 & 0.368 & 0.423 \\
 & LagNet & \best{0.309} & \best{0.289} & \best{0.269} & \best{0.261} & \best{0.246} & \best{0.275} \\
\cmidrule{1-8}
\multirow{3}{*}{Electricity} & Linear & 0.309 & 0.238 & 0.218 & 0.215 & 0.215 & 0.239 \\
 & LagLayer & 0.268 & 0.230 & 0.221 & 0.219 & 0.219 & 0.232 \\
 & LagNet & \best{0.235} & \best{0.209} & \best{0.199} & \best{0.193} & \best{0.191} & \best{0.206} \\
\cmidrule{1-8}
\multirow{3}{*}{Traffic} & Linear & 0.949 & 0.701 & 0.632 & 0.625 & 0.624 & 0.706 \\
 & LagLayer & 0.790 & 0.680 & 0.647 & 0.635 & 0.632 & 0.677 \\
 & LagNet & \best{0.606} & \best{0.505} & \best{0.477} & \best{0.465} & \best{0.457} & \best{0.502} \\
\bottomrule
\end{tabular}
}
\caption{Four-horizon average MSE under reduced training data. ETT (Avg.) aggregates the four ETT datasets. The best result in each column is shown in bold.}
\label{tab:sample-efficiency}
\end{table}

With 1\% of the training data, \method{} obtains lower MSE than
Linear on all eight datasets.
Across all training fractions, it achieves or ties for the lowest MSE
in 21 of the 25 settings.
The \laglayer{} also improves or matches Linear on four of
the five groups, supporting the effectiveness of its relative-lag
structural constraint.
It is less effective on Solar, where the additional nonlinear capacity
of \method{} produces a clear improvement.
Overall, the proposed models retain their effectiveness as the amount
of training data decreases, with \method{} showing the strongest
sample efficiency on the larger multivariate datasets.

\subsection{Forecasting Operator Analysis}

\paragraph{Forecasting matrix visualization.}
Figure~\ref{fig:matrix-visualization} compares the signed effective
matrices learned by Linear and four lag-structured variants.
The Linear matrix on ETTh1 contains prominent repeated diagonal ridges,
while Electricity and Traffic additionally exhibit variation across
historical positions and forecast steps, including vertically
concentrated responses at selected Electricity history positions and
horizontal bands in the longer-horizon cases.
These patterns suggest that relative lag provides a common organization,
but exact sharing along each diagonal is too restrictive.

The remaining columns show how the proposed parameterization represents
this structure.
The unscaled variant enforces a shared coefficient along each diagonal.
The horizon-only and history-only variants then introduce variation
across matrix rows and columns, respectively.
The complete \laglayer{} combines both forms of adaptation while
retaining the diagonal lag structure.
Rather than reproducing every Linear coefficient independently, it
represents the observed matrices through a shared lag backbone and two
lower-dimensional scale vectors.
Complete comparisons are provided in the supplementary material.

\begin{figure}[!t]
\centering
\includegraphics[width=0.98\columnwidth,draft=false]{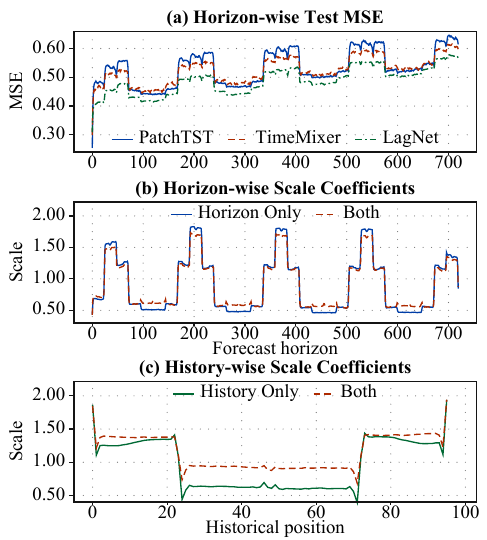}
\caption{Forecast-step MSE and learned adaptive scales on Traffic with $H=720$. Errors are averaged over test windows and channels.}
\label{fig:horizon-errors-and-scales}
\end{figure}

\paragraph{Forecast-step errors and adaptive scaling.}
Standard LTSF evaluation typically reports MSE and MAE aggregated over all forecast steps.
Figure~\ref{fig:horizon-errors-and-scales}(a) reports the test MSE at
each of the 720 forecast steps.
All three models exhibit recurring error patterns at approximately
168-step intervals, corresponding to one week in the hourly Traffic
data.
The periodic residuals of PatchTST and TimeMixer suggest that these
temporal regularities are not fully captured.
\method{} obtains the lowest MSE at almost all steps, consistent with its relative-lag parameterization better
representing recurring dependencies.

The horizon-wise profiles in panel (b) show a similar piecewise-periodic
pattern.
Their main transitions align with shifts between intervals of higher and lower forecast error, while larger relative scales generally occur in regions where the models consistently produce higher MSE.
This correspondence suggests that horizon-wise scaling adapts the
operator rows to forecast-step difficulty.

The history-wise profiles in panel (c) cannot be compared pointwise
with forecast-step MSE because they operate on historical positions.
Both variants nevertheless emphasize the earliest and most recent
parts of the input window while assigning smaller scales to many
intermediate positions.
Thus, $\boldsymbol{\alpha}$ adapts the operator across recurring
forecast regimes, whereas $\boldsymbol{\beta}$ captures nonuniform
historical-position importance.
Similar periodic error variation and structured scale responses recur
across multiple datasets.
Together, they add two-axis adaptation while retaining the
relative-lag backbone.

\subsection{Ablation and Capacity Analysis}
\paragraph{Component ablation.}
Table~\ref{tab:ablation} isolates lag sharing, adaptive scaling, and nonlinear
composition. The complete \laglayer{} matches or improves its unscaled
counterpart on all eight datasets and improves it on seven; the one-sided
variants further show that both horizon-wise and history-wise adaptation are
useful. Adaptive scaling therefore relaxes exact sharing while preserving the
structured backbone. The additional gains from \method{}, especially on the
large datasets, support nonlinear composition and multiple hidden lag-response
maps.

\begin{table}[!t]
\centering
{\small
\setlength{\tabcolsep}{1mm}
\begin{tabular}{lccccc}
\toprule
Dataset
& \shortstack{Unscaled}
& \shortstack{Horizon\\Only}
& \shortstack{History\\Only}
& \laglayer{}
& \method{} \\
\midrule
ETT (Avg.) & 0.390 & 0.389 & 0.386 & 0.385 & 0.381 \\
Weather & 0.284 & 0.279 & 0.274 & 0.273 & 0.267 \\
Solar & 0.542 & 0.440 & 0.457 & 0.371 & 0.256 \\
Electricity & 0.229 & 0.223 & 0.224 & 0.219 & 0.191 \\
Traffic & 0.681 & 0.642 & 0.652 & 0.634 & 0.460 \\
\bottomrule
\end{tabular}
}
\caption{Four-horizon average MSE for \laglayer{} variants and \method{}.}
\label{tab:ablation}
\end{table}

\paragraph{Depth and width.}
Table~\ref{tab:capacity-sweep} examines the effects of \method{} depth
and width on Solar, Electricity, and Traffic.
The depth sweep fixes $M=8$, while the width sweep fixes $R=3$.

\begin{table}[htbp]
\centering
{\small
\setlength{\tabcolsep}{0.8mm}
\begin{tabular}{@{}lcccc@{}}
\toprule
\multicolumn{5}{c}{\textbf{(a) Depth} ($M=8$)} \\
\midrule
Dataset & $R=2$ & $R=3$ & $R=4$ & $R=5$ \\
\midrule
Solar & 0.298 & 0.266 & \best{0.256} & 0.260 \\
Electricity & 0.208 & 0.201 & \best{0.194} & \best{0.194} \\
Traffic & 0.523 & 0.481 & 0.470 & \best{0.464} \\
\bottomrule
\end{tabular}

\smallskip

\begin{tabular}{@{}lcccccc@{}}
\toprule
\multicolumn{7}{c}{\textbf{(b) Width} ($R=3$)} \\
\midrule
Dataset & $M=1$ & $M=2$ & $M=4$ & $M=8$ & $M=16$ & $M=32$ \\
\midrule
Solar & 0.318 & 0.287 & 0.279 & 0.266 & \best{0.260} & 0.267 \\
Electricity & 0.221 & 0.211 & 0.204 & 0.201 & 0.193 & \best{0.192} \\
Traffic & 0.576 & 0.522 & 0.494 & 0.481 & 0.470 & \best{0.466} \\
\bottomrule
\end{tabular}
}
\caption{\method{} capacity analysis using four-horizon average MSE. Depth varies with fixed $M=8$, and width varies with fixed $R=3$. The minimum in each row is shown in bold.}
\label{tab:capacity-sweep}
\end{table}

Increasing depth generally lowers MSE, although the gains level off after
four layers on Solar and Electricity.
At fixed depth, increasing the width from $M=1$ to its best setting reduces
MSE by 13.0\% to 19.1\% across the three datasets.
Because $M=1$ already retains the nonlinear stack, these improvements reflect
the contribution of additional hidden feature maps rather than greater depth.
The gains diminish at larger widths, with Solar slightly degrading from
$M=16$ to $M=32$, suggesting that the benefit of additional capacity
eventually saturates.
Overall, depth enables repeated nonlinear composition, while width allows the
model to represent multiple lag response patterns in parallel.

\section{Conclusion}

This paper studies the structure of history-to-future forecasting operators.
\laglayer{} builds on a lag-structured matrix that shares parameters by relative lag and adapts along the forecast and history axes.
\method{} extends this operator through nonlinear depth and multiple hidden feature maps.
Matched operator-replacement studies on three established forecasters further demonstrate that relative-lag sharing transfers across decomposition-based linear, multiscale mixing, and frequency-decomposition architectures.
Across standard benchmarks, the proposed models achieve competitive accuracy; ablations, matrix diagnostics, and limited-data experiments further support the value of relative-lag sharing.
These results establish forecasting-operator structure as a useful design axis for long-term time series forecasting.

\bibliography{references}

@inproceedings{zhou2021informer,
  title = {{Informer}: Beyond Efficient Transformer for Long Sequence Time-Series Forecasting},
  author = {Zhou, Haoyi and Zhang, Shanghang and Peng, Jieqi and Zhang, Shuai and Li, Jianxin and Xiong, Hui and Zhang, Wancai},
  booktitle = {Proceedings of the AAAI Conference on Artificial Intelligence},
  pages = {11106--11115},
  doi = {10.1609/aaai.v35i12.17325},
  url = {https://ojs.aaai.org/index.php/AAAI/article/view/17325},
  year = {2021}
}

@inproceedings{wu2021autoformer,
  title = {{Autoformer}: Decomposition Transformers with Auto-Correlation for Long-Term Series Forecasting},
  author = {Wu, Haixu and Xu, Jiehui and Wang, Jianmin and Long, Mingsheng},
  booktitle = {Advances in Neural Information Processing Systems},
  volume = {34},
  pages = {22419--22430},
  url = {https://proceedings.neurips.cc/paper/2021/hash/bcc0d400288793e8bdcd7c19a8ac0c2b-Abstract.html},
  year = {2021}
}

@inproceedings{zhou2022fedformer,
  title = {{FEDformer}: Frequency Enhanced Decomposed Transformer for Long-term Series Forecasting},
  author = {Zhou, Tian and Ma, Ziqing and Wen, Qingsong and Wang, Xue and Sun, Liang and Jin, Rong},
  booktitle = {Proceedings of the 39th International Conference on Machine Learning},
  series = {Proceedings of Machine Learning Research},
  volume = {162},
  pages = {27268--27286},
  publisher = {PMLR},
  url = {https://proceedings.mlr.press/v162/zhou22g.html},
  year = {2022}
}

@inproceedings{zeng2023dlinear,
  title = {Are Transformers Effective for Time Series Forecasting?},
  author = {Zeng, Ailing and Chen, Muxi and Zhang, Lei and Xu, Qiang},
  booktitle = {Proceedings of the AAAI Conference on Artificial Intelligence},
  pages = {11121--11128},
  doi = {10.1609/aaai.v37i9.26317},
  url = {https://ojs.aaai.org/index.php/AAAI/article/view/26317},
  year = {2023}
}

@inproceedings{wu2023timesnet,
  title = {{TimesNet}: Temporal 2D-Variation Modeling for General Time Series Analysis},
  author = {Wu, Haixu and Hu, Tengge and Liu, Yong and Zhou, Hang and Wang, Jianmin and Long, Mingsheng},
  booktitle = {International Conference on Learning Representations},
  year = {2023}
}

@inproceedings{nie2023patchtst,
  title = {A Time Series is Worth 64 Words: Long-Term Forecasting with Transformers},
  author = {Nie, Yuqi and Nguyen, Nam H. and Sinthong, Phanwadee and Kalagnanam, Jayant},
  booktitle = {International Conference on Learning Representations},
  year = {2023}
}

@inproceedings{liu2024itransformer,
  title = {{iTransformer}: Inverted Transformers Are Effective for Time Series Forecasting},
  author = {Liu, Yong and Hu, Tengge and Zhang, Haoran and Wu, Haixu and Wang, Shiyu and Ma, Lintao and Long, Mingsheng},
  booktitle = {International Conference on Learning Representations},
  year = {2024}
}

@inproceedings{wang2024timemixer,
  title = {{TimeMixer}: Decomposable Multiscale Mixing for Time Series Forecasting},
  author = {Wang, Shiyu and Wu, Haixu and Shi, Xiaoming and Hu, Tengge and Luo, Huakun and Ma, Lintao and Zhang, James Y. and Zhou, Jun},
  booktitle = {International Conference on Learning Representations},
  year = {2024}
}

@article{chen2023tsmixer,
  title = {{TSMixer}: An All-{MLP} Architecture for Time Series Forecasting},
  author = {Chen, Si-An and Li, Chun-Liang and Yoder, Nathanael C. and Ar{\i}k, Sercan {\"O}. and Pfister, Tomas},
  journal = {Transactions on Machine Learning Research},
  url = {https://openreview.net/forum?id=wbpxTuXgm0},
  year = {2023}
}

@inproceedings{ekambaram2023tsmixer,
  title = {{TSMixer}: Lightweight {MLP}-Mixer Model for Multivariate Time Series Forecasting},
  author = {Ekambaram, Vijay and Jati, Arindam and Nguyen, Nam and Sinthong, Phanwadee and Kalagnanam, Jayant},
  booktitle = {Proceedings of the 29th ACM SIGKDD Conference on Knowledge Discovery and Data Mining},
  pages = {459--469},
  doi = {10.1145/3580305.3599533},
  year = {2023}
}

@inproceedings{chen2025simpletm,
  title = {{SimpleTM}: A Simple Baseline for Multivariate Time Series Forecasting},
  author = {Chen, Hui and Luong, Viet and Mukherjee, Lopamudra and Singh, Vikas},
  booktitle = {International Conference on Learning Representations},
  year = {2025}
}

@inproceedings{xu2024fits,
  title = {{FITS}: Modeling Time Series with {10k} Parameters},
  author = {Xu, Zhijian and Zeng, Ailing and Xu, Qiang},
  booktitle = {International Conference on Learning Representations},
  year = {2024}
}

@inproceedings{ma2026mixlinear,
  title = {{MixLinear}: Extreme Low Resource Multivariate Time Series Forecasting with 0.1K Parameters},
  author = {Ma, Aitian and Luo, Dongsheng and Sha, Mo},
  booktitle = {International Conference on Learning Representations},
  year = {2026}
}

@inproceedings{lin2024sparsetsf,
  title = {{SparseTSF}: Modeling Long-Term Time Series Forecasting with {1k} Parameters},
  author = {Lin, Shengsheng and Lin, Weiwei and Wu, Wentai and Chen, Haojun and Yang, Junjie},
  booktitle = {Proceedings of the 41st International Conference on Machine Learning},
  series = {Proceedings of Machine Learning Research},
  volume = {235},
  pages = {30211--30226},
  publisher = {PMLR},
  year = {2024}
}

@inproceedings{liu2022scinet,
  title = {{SCINet}: Time Series Modeling and Forecasting with Sample Convolution and Interaction},
  author = {Liu, Minhao and Zeng, Ailing and Chen, Muxi and Xu, Zhijian and Lai, Qiuxia and Ma, Lingna and Xu, Qiang},
  booktitle = {Advances in Neural Information Processing Systems},
  volume = {35},
  pages = {5816--5828},
  doi = {10.52202/068431-0421},
  url = {https://proceedings.neurips.cc/paper_files/paper/2022/hash/266983d0949aed78a16fa4782237dea7-Abstract-Conference.html},
  year = {2022}
}

@misc{bai2018tcn,
  title = {An Empirical Evaluation of Generic Convolutional and Recurrent Networks for Sequence Modeling},
  author = {Bai, Shaojie and Kolter, J. Zico and Koltun, Vladlen},
  year = {2018},
  eprint = {1803.01271},
  archivePrefix = {arXiv},
  primaryClass = {cs.LG}
}

@inproceedings{qin2023tnn,
  title = {Toeplitz Neural Network for Sequence Modeling},
  author = {Qin, Zhen and Han, Xiaodong and Sun, Weixuan and He, Bowen and Li, Dong and Li, Dongxu and Dai, Yuchao and Kong, Lingpeng and Zhong, Yiran},
  booktitle = {International Conference on Learning Representations},
  year = {2023},
  url = {https://openreview.net/forum?id=IxmWsm4xrua}
}

@inproceedings{simonyan2014saliency,
  title = {Deep Inside Convolutional Networks: Visualising Image Classification Models and Saliency Maps},
  author = {Simonyan, Karen and Vedaldi, Andrea and Zisserman, Andrew},
  booktitle = {International Conference on Learning Representations Workshop},
  year = {2014}
}

@inproceedings{ismail2020benchmarking,
  title = {Benchmarking Deep Learning Interpretability in Time Series Predictions},
  author = {Ismail, Aya Abdelsalam and Gunady, Mohamed and Corrada Bravo, H\'{e}ctor and Feizi, Soheil},
  booktitle = {Advances in Neural Information Processing Systems},
  volume = {33},
  pages = {6441--6452},
  year = {2020}
}

@article{gray2006toeplitz,
  title = {Toeplitz and Circulant Matrices: A Review},
  author = {Gray, Robert M.},
  journal = {Foundations and Trends in Communications and Information Theory},
  volume = {2},
  number = {3},
  pages = {155--239},
  doi = {10.1561/0100000006},
  url = {https://www.nowpublishers.com/article/DownloadSummary/CIT-006},
  year = {2006}
}

@inproceedings{oreshkin2020nbeats,
  title = {{N-BEATS}: Neural Basis Expansion Analysis for Interpretable Time Series Forecasting},
  author = {Oreshkin, Boris N. and Carpov, Dmitri and Chapados, Nicolas and Bengio, Yoshua},
  booktitle = {International Conference on Learning Representations},
  year = {2020}
}

@inproceedings{challu2023nhits,
  title = {{NHITS}: Neural Hierarchical Interpolation for Time Series Forecasting},
  author = {Challu, Cristian and Olivares, Kin G. and Oreshkin, Boris N. and Garza Ramirez, Federico and Mergenthaler-Canseco, Max and Dubrawski, Artur},
  booktitle = {Proceedings of the AAAI Conference on Artificial Intelligence},
  pages = {6989--6997},
  doi = {10.1609/aaai.v37i6.25854},
  url = {https://ojs.aaai.org/index.php/AAAI/article/view/25854},
  year = {2023}
}

@article{das2023tide,
  title = {Long-Term Forecasting with {TiDE}: Time-Series Dense Encoder},
  author = {Das, Abhimanyu and Kong, Weihao and Leach, Andrew and Mathur, Shaan K. and Sen, Rajat and Yu, Rose},
  journal = {Transactions on Machine Learning Research},
  year = {2023}
}

@inproceedings{luo2024moderntcn,
  title = {{ModernTCN}: A Modern Pure Convolution Structure for General Time Series Analysis},
  author = {Luo, Donghao and Wang, Xue},
  booktitle = {International Conference on Learning Representations},
  year = {2024}
}

@inproceedings{gu2022s4,
  title = {Efficiently Modeling Long Sequences with Structured State Spaces},
  author = {Gu, Albert and Goel, Karan and R{\'e}, Christopher},
  booktitle = {International Conference on Learning Representations},
  year = {2022}
}

@inproceedings{huang2025timekan,
  title = {{TimeKAN}: {KAN}-based Frequency Decomposition Learning Architecture for Long-term Time Series Forecasting},
  author = {Huang, Songtao and Zhao, Zhen and Li, Can and Bai, Lei},
  booktitle = {International Conference on Learning Representations},
  url = {https://proceedings.iclr.cc/paper_files/paper/2025/hash/46362971bfc3a97e6a271f2eb90fba17-Abstract-Conference.html},
  year = {2025}
}


\end{document}


\maketitle

\appendix

\section{Additional Method Details}
\label{app:laglayer-details}

This document provides the element-level derivation of the proposed operators,
complete reproducibility details, horizon-wise forecasting results, full
ablations, and additional structural diagnostics. Section~\ref{app:laglayer-details}
expands the single-layer operator, Section~\ref{app:deep-lagnet} details its
deep extension, and Section~\ref{app:experimental-settings} specifies the
experimental protocol used by the remaining sections.

\subsection{Indexing Conventions}
The main paper uses one chronological indexing convention throughout.
The historical input is indexed by $\ell=1,\ldots,L$, where $\ell=1$ denotes the oldest observation and $\ell=L$ denotes the most recent observation.
The forecast steps are indexed by $h=1,\ldots,H$.
The relative temporal lag between these positions is therefore $L+h-\ell$.
Accordingly, $k_1$ represents the nearest history-to-future relation, while $k_{L+H-1}$ represents the farthest relation covered by the two windows.

\subsection{Full Matrix Form}
For history length $L$ and forecast horizon $H$, the dense forecasting matrix is
\begin{equation}
\boldsymbol{W}_{\mathrm{dense}} =
\begin{bmatrix}
w_{1,1} & w_{1,2} & \cdots & w_{1,L} \\
w_{2,1} & w_{2,2} & \cdots & w_{2,L} \\
\vdots & \vdots & \ddots & \vdots \\
w_{H,1} & w_{H,2} & \cdots & w_{H,L}
\end{bmatrix},
\end{equation}
where all entries are independently learned.
The lag-structured matrix replaces these $H\times L$ independently parameterized entries with a lag kernel:
\begin{equation}
\boldsymbol{W}_{\mathrm{lag}} =
\begin{bmatrix}
k_L & k_{L-1} & \cdots & k_1 \\
k_{L+1} & k_L & \cdots & k_2 \\
\vdots & \vdots & \ddots & \vdots \\
k_{L+H-1} & k_{L+H-2} & \cdots & k_H
\end{bmatrix}.
\end{equation}
Each row and each column of $\boldsymbol{W}_{\mathrm{lag}}$ is a contiguous segment of the same lag kernel, with rows traversing the segment in reverse order.
Elements on the same top-left-to-bottom-right diagonal share one parameter.
The equality pattern is the key structural constraint.
It should not be interpreted as a rank constraint.
Rectangular Toeplitz matrices can be full rank, while low-rank factorization imposes a different assumption by forcing the matrix to pass through a low-dimensional latent factorization.

\subsection{Adaptive Scaling}
Horizon-wise and history-wise scaling can be expressed compactly as
\begin{equation}
\begin{aligned}
    \boldsymbol{W}_{\alpha}
    &=
    \boldsymbol{D}_{\alpha}\boldsymbol{W}_{\mathrm{lag}}, \\
    \boldsymbol{W}_{\beta}
    &=
    \boldsymbol{W}_{\mathrm{lag}}\boldsymbol{D}_{\beta},
\end{aligned}
\end{equation}
where $\boldsymbol{D}_{\alpha}=\operatorname{diag}(\boldsymbol{\alpha})\in\mathbb{R}^{H\times H}$ and $\boldsymbol{D}_{\beta}=\operatorname{diag}(\boldsymbol{\beta})\in\mathbb{R}^{L\times L}$ are diagonal matrices generated from the horizon-wise and history-wise scale vectors.
When both are enabled,
\begin{equation}
    \boldsymbol{W}_{\mathrm{layer}}
    =
    \boldsymbol{D}_{\alpha}
    \boldsymbol{W}_{\mathrm{lag}}
    \boldsymbol{D}_{\beta}.
\end{equation}
In all cases, the lag kernel remains the backbone of the operator.
The scale vectors only modulate row-wise or column-wise magnitude.
The complete single-layer operator with both scale vectors is denoted \laglayer{}; versions with either or both vectors removed are used only as ablations.

\section{Deep LagNet Construction}
\label{app:deep-lagnet}

A one-layer \laglayer{} is a structured linear model.
When the number of layers is one, no hidden feature maps or inter-layer activations are introduced; the model degenerates to a lag-structured linear forecaster.
The nonlinear \method{} stacks $R\geq2$ \laglayer{} modules.
The main paper presents the input-to-hidden, hidden-map mixing, and hidden-to-output matrices together with a compact layer-wise recursion, with the reported hidden temporal length fixed at $H$.
Here we introduce layer-wise dimensions only to give a unified element-level expansion and cover the implementation's optional hidden temporal length $H'$.
Let $T_r$ be the output temporal length of layer $r$ and $M_r$ its number of output feature maps, with
\begin{equation}
\begin{aligned}
    T_0&=L, & T_R&=H, & T_r&=H', &&1\leq r<R,\\
    M_0&=1, & M_R&=1, & M_r&=M, &&1\leq r<R.
\end{aligned}
    \label{eq:supp-lagnet-dimensions}
\end{equation}
The reported model sets $H'=H$.
Thus, layer $r$ maps $M_{r-1}$ input feature maps of temporal length $T_{r-1}$ to $M_r$ output feature maps of temporal length $T_r$.
For the reported setting $H'=H$, this factorization aligns the ordered hidden temporal axis with the $H$ forecast steps while using the separate feature-map axis to represent multiple learned lag-response patterns.
For output feature-map index $m$, input feature-map index $n$, output temporal position $q$, and input temporal position $p$, the corresponding scalar element is
\begin{equation}
    w^{(r)}_{m,n,q,p}
    =
    \alpha^{(r)}_{m,n,q}
    \beta^{(r)}_{m,n,p}
    k^{(r)}_{m,n,T_{r-1}+q-p}.
    \label{eq:supp-lagnet-element}
\end{equation}
Here $m=1,\ldots,M_r$, $n=1,\ldots,M_{r-1}$, $q=1,\ldots,T_r$, and $p=1,\ldots,T_{r-1}$.
Accordingly, the pair-specific lag kernel has length $T_{r-1}+T_r-1$, while the row-wise and column-wise scale vectors have lengths $T_r$ and $T_{r-1}$, respectively.
The full element-level recurrence is
\begin{equation}
    z^{(r)}_{m,q}
    =
    \sigma_r\!\left(
    \sum_{n=1}^{M_{r-1}}
    \sum_{p=1}^{T_{r-1}}
    w^{(r)}_{m,n,q,p}
    z^{(r-1)}_{n,p}
    +b^{(r)}_{m,q}
    \right).
    \label{eq:supp-lagnet-recurrence}
\end{equation}
The boundary conditions are
$\boldsymbol{z}^{(0)}_1=\boldsymbol{x}_{t-L+1:t}$ and
$\widehat{\boldsymbol{x}}_{t+1:t+H}=\boldsymbol{z}^{(R)}_1$.
The hidden-layer functions $\sigma_r$ are pointwise activations, while $\sigma_R$ is the identity map because no activation follows the final layer.
Equations~\eqref{eq:supp-lagnet-element} and~\eqref{eq:supp-lagnet-recurrence} are the element-level expansion of the compact matrix form in the main paper and match the pair-specific kernel, row-wise scale, column-wise scale, and bias parameters used in the implementation.
For the reported setting $H'=H$, the first-layer row-wise and column-wise scales are precisely the horizon-wise and history-wise scale coefficients defined for \laglayer{}.
For subsequent layers, both vectors act on the rows and columns of the future-aligned hidden maps, respectively.
Choosing $H'\neq H$ changes only the temporal dimensions of the intermediate maps; the kernel construction and layer-wise recurrence remain unchanged.
The organization of feature maps is analogous to a multi-channel convolutional layer because every output map aggregates a separate kernel response from each input map. Repeated composition of linear maps and activations gives the network its MLP depth.
We avoid adding additional branches, gates, attention modules, or decomposition blocks so that the model remains focused on the lag-structured operator.

\section{Complete Experimental Settings}
\label{app:experimental-settings}

\subsection{Datasets and Chronological Splits}

We use eight public long-term forecasting benchmarks and their established
chronological protocols~\cite{wu2023timesnet,liu2024itransformer}.
Throughout the supplementary material, Solar denotes the Solar-Energy dataset
named in the main paper.
The ETT datasets use the standard 12/4/4-month train/validation/test partition.
The remaining datasets use chronological 70\%/10\%/20\% partitions.
Table~\ref{tab:supp-dataset-details} reports the exact core split sizes used by
the data loader. Validation and test arrays prepend the final $L$ observations
of the preceding split as input context; targets remain entirely inside their
respective chronological partition. All datasets use look-back length $L=96$
and prediction lengths $H\in\{96,192,336,720\}$.

\begin{table*}[t]
\centering
{\small
\setlength{\tabcolsep}{1.2mm}
\begin{tabular}{lcccccc}
\toprule
Dataset & Variates & Frequency & Total points & Train & Validation & Test \\
\midrule
ETTh1, ETTh2 & 7 & 1 hour & 17,420 & 8,640 & 2,880 & 2,880 \\
ETTm1, ETTm2 & 7 & 15 minutes & 69,680 & 34,560 & 11,520 & 11,520 \\
Weather & 21 & 10 minutes & 52,696 & 36,887 & 5,270 & 10,539 \\
Solar & 137 & 10 minutes & 52,560 & 36,792 & 5,256 & 10,512 \\
Electricity & 321 & 1 hour & 26,304 & 18,412 & 2,632 & 5,260 \\
Traffic & 862 & 1 hour & 17,544 & 12,280 & 1,756 & 3,508 \\
\bottomrule
\end{tabular}
}
\caption{Dataset statistics and chronological core split sizes. The ETT files contain observations beyond the standard 12/4/4-month benchmark interval; these extra observations are not used.}
\label{tab:supp-dataset-details}
\end{table*}

\subsection{Preprocessing, Evaluation, and Reporting}

Feature-wise standardization statistics are fitted only on the training
partition and then applied to validation and test data.
The standalone Linear, \laglayer{}, and \method{} models additionally use
RevIN without affine parameters. Training windows are shuffled, whereas
validation and test windows retain chronological order.

For a test set containing $N_{\mathrm{test}}$ windows and $C$ variables, let
$\hat{y}_{n,c,h}$ and $y_{n,c,h}$ denote the standardized prediction and target,
respectively, for window $n$, variable $c$, and forecast step $h$.
The reported metrics in standardized space are
\begin{equation}
\begin{aligned}
\mathrm{MSE}
&=\frac{1}{N_{\mathrm{test}}CH}
\sum_{n=1}^{N_{\mathrm{test}}}\sum_{c=1}^{C}\sum_{h=1}^{H}
(\hat{y}_{n,c,h}-y_{n,c,h})^2,\\
\mathrm{MAE}
&=\frac{1}{N_{\mathrm{test}}CH}
\sum_{n=1}^{N_{\mathrm{test}}}\sum_{c=1}^{C}\sum_{h=1}^{H}
|\hat{y}_{n,c,h}-y_{n,c,h}|.
\end{aligned}
\label{eq:supp-evaluation-metrics}
\end{equation}
Following standard LTSF evaluation protocols, we report both metrics because
MSE penalizes large deviations more strongly, whereas MAE provides a
complementary measure of absolute error.
The checkpoint with the lowest validation MSE is evaluated once on the test
partition. Unless a diagnostic experiment explicitly states otherwise,
reported reproduced and proposed results use seeds
$\{2027,2026,2025,2024,2023\}$. Tables report the arithmetic mean and sample
standard deviation over the five runs. Trainable parameter counts are computed
as $\sum_{\theta:\,\mathrm{requires\_grad}}|\theta|$ for the instantiated
model, including normalization and bias parameters.

\subsection{Computing Environment}

Independent runs were distributed across two Windows workstations for parallel
execution. Workstation A ran Windows 10 Enterprise LTSC (build 19044) with two
Intel Xeon Gold 6430 CPUs, 255.4~GB RAM, and four NVIDIA GeForce RTX 4090 GPUs
(24~GB each; driver 576.02). Its software environment used Python 3.11.5,
PyTorch 2.2.0 with CUDA 12.1 and cuDNN 8.8.1, NumPy 1.26.4, pandas 2.2.2, and
Matplotlib 3.8.4. Workstation B ran Windows 10 (build 19045) with an Intel Xeon
Gold 6248R CPU, 255.6~GB RAM, and one NVIDIA RTX A6000 GPU (48~GB; driver
595.95). It used Python 3.11.11, PyTorch 2.1.0 with CUDA 12.1 and cuDNN 8.8.1,
NumPy 1.26.4, pandas 2.2.3, and Matplotlib 3.10.9. Environment manifests for
both workstations are included in the code supplement.

\subsection{Standalone Lag-Structured Models}

Table~\ref{tab:supp-main-settings} gives the exact architecture, parameter
sharing mode, and batch size used for the standalone \laglayer{} and
\method{} columns. A shared operator applies one channel-independent temporal
model to every variable; a per-variable setting learns a separate temporal
model for each variable. Both proposed models use Adam with learning rate
$10^{-3}$, at most 20 epochs, early-stopping patience 5, gradient clipping at
1.0, and automatic mixed precision. The global and lag-kernel weight decays
are $10^{-5}$; scale coefficients and biases are not decayed. Adaptive scales
are initialized to one, biases are enabled, and hidden temporal length equals
$H$.
During development of the standalone lag-structured models, batch size was the
only tuned optimization hyperparameter. We tried batch sizes
$\{16,64,256\}$ and selected the dataset-specific value by validation MSE; all
other settings were fixed as reported above.

\begin{table*}[t]
\centering
{\small
\setlength{\tabcolsep}{1mm}
\begin{tabular}{lccccc}
\toprule
Dataset & Temporal operator & Batch size & \laglayer{} $(R,M)$ & \method{} $(R,M)$ & Adaptive scales \\
\midrule
ETTh1 & Shared & 256 & $(1,\text{--})$ & $(2,4)$ & Both \\
ETTh2 & Shared & 64 & $(1,\text{--})$ & $(2,4)$ & Both \\
ETTm1 & Per-variable & 256 & $(1,\text{--})$ & $(2,4)$ & Both \\
ETTm2 & Per-variable & 64 & $(1,\text{--})$ & $(2,4)$ & Both \\
Weather & Per-variable & 16 & $(1,\text{--})$ & $(2,4)$ & Both \\
Solar & Shared & 64 & $(1,\text{--})$ & $(4,12)$ & Both \\
Electricity & Shared & 64 & $(1,\text{--})$ & $(4,12)$ & Both \\
Traffic & Shared & 64 & $(1,\text{--})$ & $(4,12)$ & Both \\
\bottomrule
\end{tabular}
}
\caption{Standalone lag-structured configurations used in the main forecasting table. $R$ is the number of lag-structured layers and $M$ the hidden feature-map width. Both denotes horizon-wise and history-wise scaling; \method{} applies the corresponding row-wise and column-wise mechanisms in every layer.}
\label{tab:supp-main-settings}
\end{table*}

\subsection{Operator-Transfer Configurations}

DLinear, TimeMixer, and TimeKAN, together with their lag-structured variants, follow the
model architectures and implementation details in their official papers and
released repositories.
Their configurations are summarized in
Tables~\ref{tab:supp-dlinear-settings}, \ref{tab:supp-timemixer-settings},
and~\ref{tab:supp-timekan-settings}.
For a controlled comparison, each reproduced model and its lag-structured
variant use identical dataset-specific training settings.
Within each pair, the data pipeline, optimization settings, random seed, and
all components outside the replaced temporal mapping remain identical.
DLagLayer preserves the moving-average decomposition and branch summation,
replacing only the independent seasonal and trend
$\operatorname{Linear}(L,H)$ projections with dimension-matched
\laglayer{} operators.
Lag-TimeMixer replaces each FMM temporal predictor
$\operatorname{Linear}(T_i,H)$ with a dimension-matched \laglayer{}.
Lag-TimeKAN preserves TimeKAN's frequency decomposition, M-KAN blocks,
depthwise convolutions, normalization, and output projection, replacing only
the final $\operatorname{Linear}(L,H)$ temporal predictor with a
dimension-matched \laglayer{} using both adaptive scales.

\begin{table*}[t]
\centering
{\small
\setlength{\tabcolsep}{1.2mm}
\begin{tabular}{lcccccccc}
\toprule
Dataset & Batch size & $\eta_{96}$ & $\eta_{192}$ & $\eta_{336}$ & $\eta_{720}$ & Epochs & Patience & Individual \\
\midrule
ETTh1       & 32 & $5{\times}10^{-3}$ & $5{\times}10^{-3}$ & $5{\times}10^{-3}$ & $5{\times}10^{-3}$ & 10 & 3 & No \\
ETTh2       & 32 & $5{\times}10^{-2}$ & $5{\times}10^{-2}$ & $5{\times}10^{-2}$ & $5{\times}10^{-2}$ & 10 & 3 & No \\
ETTm1       & 8  & $1{\times}10^{-4}$ & $1{\times}10^{-4}$ & $1{\times}10^{-4}$ & $1{\times}10^{-4}$ & 10 & 3 & No \\
ETTm2       & 32 & $1{\times}10^{-3}$ & $1{\times}10^{-3}$ & $1{\times}10^{-2}$ & $1{\times}10^{-2}$ & 10 & 3 & No \\
Weather     & 16 & $1{\times}10^{-4}$ & $1{\times}10^{-4}$ & $1{\times}10^{-4}$ & $1{\times}10^{-4}$ & 10 & 3 & No \\
Solar       & 16 & $1{\times}10^{-3}$ & $1{\times}10^{-3}$ & $1{\times}10^{-3}$ & $1{\times}10^{-3}$ & 10 & 3 & No \\
Electricity & 16 & $1{\times}10^{-3}$ & $1{\times}10^{-3}$ & $1{\times}10^{-3}$ & $1{\times}10^{-3}$ & 10 & 3 & No \\
Traffic     & 16 & $5{\times}10^{-2}$ & $5{\times}10^{-2}$ & $5{\times}10^{-2}$ & $5{\times}10^{-2}$ & 10 & 3 & No \\
\bottomrule
\end{tabular}
}
\caption{DLinear and DLagLayer configurations. Here $\eta_H$ denotes the
learning rate at forecast horizon $H$. The shared-variable DLinear
architecture, moving-average kernel 25, MSE loss, Adam optimizer, official
type-1 learning-rate schedule, and dataset-specific settings follow the
official paper and repository. All runs use $L=96$. Solar is absent from the
official DLinear benchmark scripts and therefore uses the complete official
Electricity training setting.}
\label{tab:supp-dlinear-settings}
\end{table*}

\begin{table*}[t]
\centering
{\small
\setlength{\tabcolsep}{1mm}
\begin{tabular}{lccccccccc}
\toprule
Dataset & $M$ & PDM blocks & $d_{\mathrm{model}}$ & $d_{\mathrm{ff}}$ & Batch size & Epochs & Patience & Channel independent & Normalize \\
\midrule
ETTh1       & 3 & 2 & 16  & 32   & 128 & 10 & 10 & Yes & Yes \\
ETTh2       & 3 & 2 & 16  & 32   & 128 & 10 & 10 & Yes & Yes \\
ETTm1       & 3 & 2 & 16  & 32   & 128 & 10 & 10 & Yes & Yes \\
ETTm2       & 3 & 2 & 32  & 32   & 128 & 10 & 10 & Yes & Yes \\
Weather     & 3 & 2 & 16  & 32   & 128 & 20 & 10 & Yes & Yes \\
Solar       & 3 & 2 & 128 & 2048 & 32  & 20 & 3  & No  & No \\
Electricity & 3 & 2 & 16  & 32   & 32  & 20 & 10 & Yes & Yes \\
Traffic     & 3 & 2 & 32  & 64   & 8   & 20 & 15 & Yes & Yes \\
\bottomrule
\end{tabular}
}
\caption{TimeMixer and Lag-TimeMixer configurations used in the main forecasting table. The core settings $M$, number of PDM blocks, $d_{\mathrm{model}}$, learning rate, loss, batch size, and epochs follow Table~7 of the official TimeMixer paper; unreported model and optimization options are taken from the released implementation. All configurations use $L=96$, four prediction lengths, MSE loss, Adam with learning rate $10^{-2}$, OneCycleLR with \texttt{pct\_start}=0.2, average-pooling downsampling with window 2, moving-average decomposition with kernel 25, and dropout 0.1. Here $M=3$ denotes three downsampling stages and hence four temporal scales.}
\label{tab:supp-timemixer-settings}
\end{table*}

\begin{table*}[t]
\centering
{\small
\setlength{\tabcolsep}{1.2mm}
\begin{tabular}{lccccccc}
\toprule
Dataset & $H=96$ & $H=192$ & $H=336$ & $H=720$
& Batch size & Epochs & Patience \\
\midrule
ETTh1       & $(16,2,2,0)$ & $(16,2,1,0)$ & $(16,2,2,0)$ & $(16,2,3,1)$ & 128 & 10 & 10 \\
ETTh2       & $(16,2,3,0)$ & $(16,1,1,0)$ & $(32,1,3,0)$ & $(16,2,3,1)$ & 128 & 10 & 10 \\
ETTm1       & $(16,2,2,1)$ & $(16,2,2,1)$ & $(16,3,2,1)$ & $(16,2,3,1)$ & 128 & 10 & 10 \\
ETTm2       & $(32,3,2,1)$ & $(32,3,2,1)$ & $(32,1,1,1)$ & $(16,3,3,1)$ & 128 & 10 & 10 \\
Weather     & $(16,3,3,1)$ & $(16,3,2,1)$ & $(16,3,2,1)$ & $(16,3,2,1)$ & 128 & 20 & 10 \\
Solar       & $(16,3,3,1)$ & $(16,3,3,1)$ & $(16,3,3,1)$ & $(16,3,3,1)$ & 32 & 20 & 3 \\
Electricity & $(16,3,3,1)$ & $(16,3,3,1)$ & $(16,3,3,1)$ & $(16,3,3,1)$ & 32 & 20 & 10 \\
Traffic     & $(16,3,3,1)$ & $(16,3,3,1)$ & $(16,3,3,1)$ & $(16,3,3,1)$ & 8 & 20 & 15 \\
\bottomrule
\end{tabular}
}
\caption{TimeKAN and Lag-TimeKAN configurations. Each horizon-specific tuple is $(d_{\mathrm{model}},E,D,O)$, where $E$ is the number of encoder layers, $D$ the number of downsampling layers, and $O$ the beginning decomposition order. The model configurations follow the official TimeKAN paper and released main-experiment scripts; Solar and Traffic use the released Electricity configuration as a fixed extension. All runs use $L=96$, $d_{\mathrm{ff}}=32$, downsampling window 2, moving-average kernel 25, dropout 0.1, normalization, MSE loss, Adam with learning rate $10^{-2}$, and OneCycleLR with \texttt{pct\_start}=0.2. The dataset-specific batch sizes, epoch limits, and patience values form the unified training protocol shared by TimeKAN and Lag-TimeKAN.}
\label{tab:supp-timekan-settings}
\end{table*}

\begin{table*}[t]
\centering
{\small
\setlength{\tabcolsep}{1mm}
\begin{tabular}{ll*{5}{cc}}
\toprule
Dataset & $H$ & \multicolumn{2}{c}{iTransformer} & \multicolumn{2}{c}{DLinear} & \multicolumn{2}{c}{PatchTST} & \multicolumn{2}{c}{TimeMixer} & \multicolumn{2}{c}{TimeKAN} \\
\cmidrule(lr){3-4}\cmidrule(lr){5-6}\cmidrule(lr){7-8}\cmidrule(lr){9-10}\cmidrule(lr){11-12}
& & MSE & MAE & MSE & MAE & MSE & MAE & MSE & MAE & MSE & MAE \\
\midrule
ETTh1 & 96 & 0.386 & 0.405 & 0.390 & 0.404 & 0.460 & 0.447 & 0.380 & 0.401 & 0.370 & 0.397 \\
 & 192 & 0.441 & 0.436 & 0.448 & 0.443 & 0.512 & 0.477 & 0.446 & 0.434 & 0.416 & 0.421 \\
 & 336 & 0.487 & 0.458 & 0.499 & 0.475 & 0.546 & 0.496 & 0.490 & 0.455 & 0.446 & 0.437 \\
 & 720 & 0.503 & 0.491 & 0.515 & 0.509 & 0.544 & 0.517 & 0.538 & 0.500 & 0.456 & 0.460 \\
\midrule
ETTh2 & 96 & 0.297 & 0.349 & 0.374 & 0.410 & 0.308 & 0.355 & 0.297 & 0.348 & 0.296 & 0.345 \\
 & 192 & 0.380 & 0.400 & 0.468 & 0.465 & 0.393 & 0.405 & 0.380 & 0.400 & 0.380 & 0.396 \\
 & 336 & 0.428 & 0.432 & 0.510 & 0.495 & 0.427 & 0.436 & 0.402 & 0.424 & 0.396 & 0.419 \\
 & 720 & 0.427 & 0.445 & 0.693 & 0.596 & 0.436 & 0.450 & 0.423 & 0.443 & 0.431 & 0.444 \\
\midrule
ETTm1 & 96 & 0.334 & 0.368 & 0.345 & 0.372 & 0.352 & 0.374 & 0.321 & 0.359 & 0.330 & 0.367 \\
 & 192 & 0.377 & 0.391 & 0.383 & 0.392 & 0.390 & 0.393 & 0.363 & 0.383 & 0.359 & 0.384 \\
 & 336 & 0.426 & 0.420 & 0.415 & 0.416 & 0.421 & 0.414 & 0.395 & 0.407 & 0.388 & 0.405 \\
 & 720 & 0.491 & 0.459 & 0.476 & 0.455 & 0.462 & 0.449 & 0.456 & 0.444 & 0.448 & 0.439 \\
\midrule
ETTm2 & 96 & 0.180 & 0.264 & 0.188 & 0.284 & 0.183 & 0.270 & 0.177 & 0.259 & 0.175 & 0.257 \\
 & 192 & 0.250 & 0.309 & 0.270 & 0.342 & 0.255 & 0.314 & 0.242 & 0.302 & 0.243 & 0.305 \\
 & 336 & 0.311 & 0.348 & 0.346 & 0.393 & 0.309 & 0.347 & 0.298 & 0.339 & 0.305 & 0.344 \\
 & 720 & 0.412 & 0.407 & 0.480 & 0.467 & 0.412 & 0.404 & 0.398 & 0.399 & 0.413 & 0.411 \\
\midrule
Weather & 96 & 0.174 & 0.214 & 0.197 & 0.257 & 0.186 & 0.227 & 0.164 & 0.211 & 0.164 & 0.211 \\
 & 192 & 0.221 & 0.254 & 0.238 & 0.295 & 0.234 & 0.265 & 0.211 & 0.253 & 0.208 & 0.251 \\
 & 336 & 0.278 & 0.296 & 0.284 & 0.333 & 0.284 & 0.301 & 0.265 & 0.292 & 0.263 & 0.291 \\
 & 720 & 0.358 & 0.347 & 0.348 & 0.384 & 0.356 & 0.349 & 0.343 & 0.346 & 0.339 & 0.343 \\
\midrule
Solar & 96 & 0.203 & 0.237 & 0.285 & 0.372 & 0.265 & 0.323 & 0.210 & 0.284 & 0.239 & 0.287 \\
 & 192 & 0.233 & 0.261 & 0.316 & 0.394 & 0.288 & 0.332 & 0.225 & 0.300 & 0.279 & 0.316 \\
 & 336 & 0.248 & 0.273 & 0.349 & 0.412 & 0.301 & 0.339 & 0.243 & 0.303 & 0.310 & 0.334 \\
 & 720 & 0.249 & 0.275 & 0.354 & 0.410 & 0.295 & 0.336 & 0.244 & 0.304 & 0.300 & 0.322 \\
\midrule
Electricity & 96 & 0.148 & 0.240 & 0.195 & 0.278 & 0.190 & 0.296 & 0.155 & 0.246 & 0.176 & 0.269 \\
 & 192 & 0.162 & 0.253 & 0.194 & 0.280 & 0.199 & 0.304 & 0.167 & 0.258 & 0.182 & 0.273 \\
 & 336 & 0.178 & 0.269 & 0.207 & 0.296 & 0.217 & 0.319 & 0.184 & 0.275 & 0.198 & 0.289 \\
 & 720 & 0.225 & 0.317 & 0.242 & 0.329 & 0.258 & 0.352 & 0.228 & 0.314 & 0.241 & 0.324 \\
\midrule
Traffic & 96 & 0.395 & 0.268 & 0.649 & 0.397 & 0.526 & 0.347 & 0.463 & 0.289 & 0.585 & 0.380 \\
 & 192 & 0.417 & 0.276 & 0.599 & 0.371 & 0.522 & 0.332 & 0.482 & 0.299 & 0.559 & 0.363 \\
 & 336 & 0.433 & 0.283 & 0.606 & 0.374 & 0.517 & 0.334 & 0.493 & 0.304 & 0.571 & 0.364 \\
 & 720 & 0.467 & 0.302 & 0.646 & 0.396 & 0.552 & 0.352 & 0.525 & 0.321 & 0.603 & 0.381 \\
\bottomrule
\end{tabular}
}
\caption{Horizon-wise baseline results for $L=96$. The PatchTST values are copied from the unified TimeMixer evaluation, and iTransformer values from its official full forecasting table. DLinear, TimeMixer, and TimeKAN entries are five-seed means from our reproductions under the corresponding protocols. Each entry reports MSE/MAE.}
\label{tab:supp-public-full-results}
\end{table*}

\section{Complete Forecasting Results}
\label{app:complete-results}

The main paper reports four-horizon average MSE and MAE together with a compact paired operator-replacement comparison.
The tables below provide the corresponding horizon-wise metrics, seed statistics, and parameter audits.
All reproduced and proposed results are averaged over five random seeds.

\subsection{Public Baselines and Parameter Audit}

Table~\ref{tab:supp-public-full-results} gives the full public baseline results.
The PatchTST column is copied from the unified $L=96$ appendix of TimeMixer~\cite{wang2024timemixer}, while the iTransformer column is copied from the official $L=96$ full forecasting table of iTransformer~\cite{liu2024itransformer}.
For TimeMixer, we reproduce all eight datasets under the main-experiment configurations reported in Table~7 of the official paper, using the released implementation to determine model and optimization options not specified by that table, and report the mean over the same five seeds as our methods.
This additional reproduction is used because the published results do not provide statistics under the same protocol with five random seeds.
For TimeKAN, we preserve the official model architecture and
horizon-specific configurations while applying the unified dataset-specific
training protocol in Table~\ref{tab:supp-timekan-settings} to the same five
random seeds.
Table~\ref{tab:supp-public-param-counts} reports the audited public-baseline parameter counts used in the main table.

\begin{table*}[t]
\centering
{\small
\setlength{\tabcolsep}{1mm}
\begin{tabular}{llccccc}
\toprule
Dataset & $H$ & iTransformer & DLinear & PatchTST & TimeMixer & TimeKAN \\
\midrule
ETTh1 & 96 & 841.6K & 18.6K & 35.2K & 75.4K & 12.8K \\
 & 192 & 866.2K & 37.2K & 53.7K & 93.1K & 20.5K \\
 & 336 & 3.38M & 65.2K & 81.5K & 119.6K & 36.1K \\
 & 720 & 3.58M & 139.7K & 155.6K & 190.2K & 77.6K \\
\midrule
ETTh2 & 96 & 224.2K & 18.6K & 35.2K & 75.4K & 15.0K \\
 & 192 & 236.6K & 37.2K & 53.7K & 93.1K & 19.6K \\
 & 336 & 255.2K & 65.2K & 81.5K & 119.6K & 43.5K \\
 & 720 & 304.7K & 139.7K & 155.6K & 190.2K & 77.6K \\
\midrule
ETTm1 & 96 & 224.2K & 18.6K & 548.7K & 75.4K & 14.4K \\
 & 192 & 236.6K & 37.2K & 696.3K & 93.1K & 23.7K \\
 & 336 & 255.2K & 65.2K & 917.6K & 119.6K & 40.1K \\
 & 720 & 304.7K & 139.7K & 1.51M & 190.2K & 77.6K \\
\midrule
ETTm2 & 96 & 224.2K & 18.6K & 548.7K & 77.6K & 38.1K \\
 & 192 & 236.6K & 37.2K & 696.3K & 95.3K & 47.4K \\
 & 336 & 255.2K & 65.2K & 917.6K & 121.8K & 38.2K \\
 & 720 & 304.7K & 139.7K & 1.51M & 192.5K & 81.4K \\
\midrule
Weather & 96 & 4.83M & 18.6K & 548.7K & 75.5K & 20.9K \\
 & 192 & 4.88M & 37.2K & 696.3K & 93.2K & 26.2K \\
 & 336 & 4.96M & 65.2K & 917.6K & 119.7K & 40.2K \\
 & 720 & 5.15M & 139.7K & 1.51M & 190.4K & 77.4K \\
\midrule
Solar & 96 & 3.26M & 18.6K & 548.7K & 2.28M & 21.9K \\
 & 192 & 3.31M & 37.2K & 696.3K & 2.32M & 31.2K \\
 & 336 & 3.38M & 65.2K & 917.6K & 2.37M & 45.1K \\
 & 720 & 3.58M & 139.7K & 1.51M & 2.51M & 82.4K \\
\midrule
Electricity & 96 & 4.83M & 18.6K & 548.7K & 77.9K & 23.3K \\
 & 192 & 4.88M & 37.2K & 696.3K & 95.6K & 32.6K \\
 & 336 & 4.96M & 65.2K & 917.6K & 122.1K & 46.6K \\
 & 720 & 5.15M & 139.7K & 1.51M & 192.8K & 83.9K \\
\midrule
Traffic & 96 & 6.41M & 18.6K & 548.7K & 88.6K & 27.7K \\
 & 192 & 6.46M & 37.2K & 696.3K & 106.3K & 37.0K \\
 & 336 & 6.53M & 65.2K & 917.6K & 132.8K & 50.9K \\
 & 720 & 6.73M & 139.7K & 1.51M & 203.5K & 88.2K \\
\bottomrule
\end{tabular}
}
\caption{Audited horizon-wise trainable parameter counts for the public baselines. Counts use the model definitions and configurations underlying the reported $L=96$ results; DLinear, TimeMixer, and TimeKAN counts come from our reproduced implementations.}
\label{tab:supp-public-param-counts}
\end{table*}

\subsection{Operator-Transfer Results}

Table~\ref{tab:supp-dlinear-pair} reports the paired DLinear and DLagLayer
results.
Both variants follow the settings in Table~\ref{tab:supp-dlinear-settings};
DLagLayer changes only the seasonal and trend temporal projections.
Table~\ref{tab:supp-timemixer-pair} reports the paired TimeMixer and Lag-TimeMixer seed statistics.
Lag-TimeMixer preserves the complete TimeMixer PDM feature extractor and multiscale aggregation and follows the same paired training protocol, replacing only each FMM temporal predictor $\operatorname{Linear}(T_i,H)$ with a \laglayer{} operator of the same input and output lengths.
Table~\ref{tab:supp-timekan-lag-results} reports the corresponding TimeKAN and
Lag-TimeKAN results and run counts.
Lag-TimeKAN preserves TimeKAN's complete frequency-decomposition feature
extractor and replaces only its final linear temporal predictor with the
complete \laglayer{} operator.
For the TimeMixer and TimeKAN paired studies, all 32 dataset--horizon settings
contain complete records for the same five random seeds in each model.
The paired operator-replacement table in the main paper uses four-horizon averages.
The DLinear--DLagLayer, TimeMixer--Lag-TimeMixer, and
TimeKAN--Lag-TimeKAN blocks use the corresponding paired results reported
below.
MSE change is $100(\mathrm{MSE}_{\mathrm{lag}}-\mathrm{MSE}_{\mathrm{base}})/\mathrm{MSE}_{\mathrm{base}}$.
The body of that table reports the four-horizon average trainable parameter counts before and after replacement, while its final row reports the mean relative changes in MSE and parameter count across the underlying datasets.
The final average treats each underlying dataset equally; in particular, the
four ETT datasets contribute four separate terms rather than a single
aggregated term.
These tables are separated for readability and to make the source of each number explicit.

\begin{table*}[t]
\centering
{\small
\setlength{\tabcolsep}{0.75mm}
\begin{tabular}{ll*{2}{ccc}}
\toprule
Dataset & $H$ & \multicolumn{3}{c}{DLinear} & \multicolumn{3}{c}{DLagLayer} \\
\cmidrule(lr){3-5}\cmidrule(lr){6-8}
& & Params & MSE & MAE & Params & MSE & MAE \\
\midrule
ETTh1 & 96  & 18.6K  & 0.382 & 0.393 & 958  & 0.392 & 0.407 \\
      & 192 & 37.2K  & 0.438 & 0.430 & 1.5K & 0.434 & 0.428 \\
      & 336 & 65.2K  & 0.513 & 0.488 & 2.4K & 0.482 & 0.457 \\
      & 720 & 139.7K & 0.507 & 0.502 & 4.7K & 0.505 & 0.501 \\
\midrule
ETTh2 & 96  & 18.6K  & 0.470 & 0.474 & 958  & 0.317 & 0.375 \\
      & 192 & 37.2K  & 0.449 & 0.451 & 1.5K & 0.434 & 0.446 \\
      & 336 & 65.2K  & 0.532 & 0.505 & 2.4K & 0.575 & 0.532 \\
      & 720 & 139.7K & 0.651 & 0.580 & 4.7K & 0.754 & 0.620 \\
\midrule
ETTm1 & 96  & 18.6K  & 0.345 & 0.373 & 958  & 0.350 & 0.375 \\
      & 192 & 37.2K  & 0.385 & 0.395 & 1.5K & 0.386 & 0.394 \\
      & 336 & 65.2K  & 0.414 & 0.413 & 2.4K & 0.417 & 0.417 \\
      & 720 & 139.7K & 0.474 & 0.453 & 4.7K & 0.476 & 0.454 \\
\midrule
ETTm2 & 96  & 18.6K  & 0.186 & 0.280 & 958  & 0.185 & 0.274 \\
      & 192 & 37.2K  & 0.288 & 0.360 & 1.5K & 0.275 & 0.352 \\
      & 336 & 65.2K  & 0.356 & 0.405 & 2.4K & 0.337 & 0.386 \\
      & 720 & 139.7K & 0.476 & 0.457 & 4.7K & 0.459 & 0.462 \\
\bottomrule
\end{tabular}
\hfill
\begin{tabular}{ll*{2}{ccc}}
\toprule
Dataset & $H$ & \multicolumn{3}{c}{DLinear} & \multicolumn{3}{c}{DLagLayer} \\
\cmidrule(lr){3-5}\cmidrule(lr){6-8}
& & Params & MSE & MAE & Params & MSE & MAE \\
\midrule
Weather & 96  & 18.6K  & 0.197 & 0.256 & 958  & 0.204 & 0.267 \\
        & 192 & 37.2K  & 0.238 & 0.294 & 1.5K & 0.244 & 0.303 \\
        & 336 & 65.2K  & 0.283 & 0.332 & 2.4K & 0.291 & 0.344 \\
        & 720 & 139.7K & 0.348 & 0.385 & 4.7K & 0.350 & 0.387 \\
\midrule
Solar & 96  & 18.6K  & 0.284 & 0.372 & 958  & 0.286 & 0.373 \\
      & 192 & 37.2K  & 0.316 & 0.394 & 1.5K & 0.320 & 0.387 \\
      & 336 & 65.2K  & 0.349 & 0.411 & 2.4K & 0.355 & 0.404 \\
      & 720 & 139.7K & 0.354 & 0.411 & 4.7K & 0.362 & 0.405 \\
\midrule
Electricity & 96  & 18.6K  & 0.195 & 0.278 & 958  & 0.199 & 0.282 \\
            & 192 & 37.2K  & 0.194 & 0.280 & 1.5K & 0.198 & 0.284 \\
            & 336 & 65.2K  & 0.207 & 0.296 & 2.4K & 0.210 & 0.299 \\
            & 720 & 139.7K & 0.243 & 0.329 & 4.7K & 0.246 & 0.332 \\
\midrule
Traffic & 96  & 18.6K  & 0.650 & 0.398 & 958  & 0.653 & 0.398 \\
        & 192 & 37.2K  & 0.599 & 0.371 & 1.5K & 0.605 & 0.374 \\
        & 336 & 65.2K  & 0.605 & 0.374 & 2.4K & 0.613 & 0.377 \\
        & 720 & 139.7K & 0.646 & 0.395 & 4.7K & 0.651 & 0.398 \\
\bottomrule
\end{tabular}
}
\caption{Paired DLinear and DLagLayer results. Both models use the same
official shared-variable DLinear protocol. DLagLayer preserves the
moving-average decomposition and replaces only the independent seasonal and
trend temporal projections with dimension-matched \laglayer{} operators.
Solar uses the complete official Electricity training setting because it is
absent from the official DLinear benchmark scripts. Params denotes total
trainable parameters. Results are reported from one run with random seed
2027.}
\label{tab:supp-dlinear-pair}
\end{table*}

\begin{table*}[t]
\centering
{\small
\setlength{\tabcolsep}{1mm}
\begin{tabular}{ll*{2}{cccc}}
\toprule
Dataset & $H$ & \multicolumn{4}{c}{TimeMixer} & \multicolumn{4}{c}{Lag-TimeMixer} \\
\cmidrule(lr){3-6}\cmidrule(lr){7-10}
& & Params & Runs & MSE & MAE & Params & Runs & MSE & MAE \\
\midrule
\multirow{4}{*}{ETTh1} & 96 & 75.4K & 5/5 & 0.380$\pm$0.008 & 0.401$\pm$0.002 & 59.3K & 5/5 & 0.378$\pm$0.007 & 0.397$\pm$0.004 \\
 & 192 & 93.1K & 5/5 & 0.446$\pm$0.006 & 0.434$\pm$0.003 & 60.4K & 5/5 & 0.435$\pm$0.006 & 0.428$\pm$0.002 \\
 & 336 & 119.6K & 5/5 & 0.490$\pm$0.010 & 0.455$\pm$0.009 & 62.2K & 5/5 & 0.484$\pm$0.013 & 0.448$\pm$0.007 \\
 & 720 & 190.2K & 5/5 & 0.538$\pm$0.026 & 0.500$\pm$0.015 & 66.8K & 5/5 & 0.539$\pm$0.046 & 0.501$\pm$0.023 \\
\midrule
\multirow{4}{*}{ETTh2} & 96 & 75.4K & 5/5 & 0.297$\pm$0.002 & 0.348$\pm$0.002 & 59.3K & 5/5 & 0.295$\pm$0.004 & 0.346$\pm$0.004 \\
 & 192 & 93.1K & 5/5 & 0.380$\pm$0.003 & 0.400$\pm$0.003 & 60.4K & 5/5 & 0.374$\pm$0.006 & 0.395$\pm$0.003 \\
 & 336 & 119.6K & 5/5 & 0.402$\pm$0.011 & 0.424$\pm$0.007 & 62.2K & 5/5 & 0.388$\pm$0.010 & 0.413$\pm$0.006 \\
 & 720 & 190.2K & 5/5 & 0.423$\pm$0.006 & 0.443$\pm$0.004 & 66.8K & 5/5 & 0.434$\pm$0.010 & 0.448$\pm$0.008 \\
\midrule
\multirow{4}{*}{ETTm1} & 96 & 75.4K & 5/5 & 0.321$\pm$0.007 & 0.359$\pm$0.005 & 59.3K & 5/5 & 0.321$\pm$0.003 & 0.359$\pm$0.002 \\
 & 192 & 93.1K & 5/5 & 0.363$\pm$0.003 & 0.383$\pm$0.002 & 60.4K & 5/5 & 0.363$\pm$0.003 & 0.384$\pm$0.003 \\
 & 336 & 119.6K & 5/5 & 0.395$\pm$0.003 & 0.407$\pm$0.004 & 62.2K & 5/5 & 0.393$\pm$0.003 & 0.406$\pm$0.002 \\
 & 720 & 190.2K & 5/5 & 0.456$\pm$0.004 & 0.444$\pm$0.004 & 66.8K & 5/5 & 0.454$\pm$0.003 & 0.442$\pm$0.002 \\
\midrule
\multirow{4}{*}{ETTm2} & 96 & 77.6K & 5/5 & 0.177$\pm$0.001 & 0.259$\pm$0.001 & 61.5K & 5/5 & 0.175$\pm$0.000 & 0.258$\pm$0.001 \\
 & 192 & 95.3K & 5/5 & 0.242$\pm$0.002 & 0.302$\pm$0.002 & 62.6K & 5/5 & 0.240$\pm$0.002 & 0.301$\pm$0.003 \\
 & 336 & 121.8K & 5/5 & 0.298$\pm$0.003 & 0.339$\pm$0.002 & 64.4K & 5/5 & 0.300$\pm$0.001 & 0.340$\pm$0.002 \\
 & 720 & 192.5K & 5/5 & 0.398$\pm$0.005 & 0.399$\pm$0.005 & 69.0K & 5/5 & 0.395$\pm$0.004 & 0.396$\pm$0.002 \\
\midrule
\multirow{4}{*}{Weather} & 96 & 75.5K & 5/5 & 0.164$\pm$0.001 & 0.211$\pm$0.001 & 59.4K & 5/5 & 0.163$\pm$0.001 & 0.210$\pm$0.001 \\
 & 192 & 93.2K & 5/5 & 0.211$\pm$0.001 & 0.253$\pm$0.002 & 60.5K & 5/5 & 0.211$\pm$0.001 & 0.254$\pm$0.002 \\
 & 336 & 119.7K & 5/5 & 0.265$\pm$0.002 & 0.292$\pm$0.001 & 62.3K & 5/5 & 0.265$\pm$0.001 & 0.293$\pm$0.000 \\
 & 720 & 190.4K & 5/5 & 0.343$\pm$0.004 & 0.346$\pm$0.003 & 66.9K & 5/5 & 0.342$\pm$0.002 & 0.346$\pm$0.001 \\
\midrule
\multirow{4}{*}{Solar} & 96 & 2.28M & 5/5 & 0.210$\pm$0.011 & 0.284$\pm$0.004 & 2.25M & 5/5 & 0.202$\pm$0.010 & 0.289$\pm$0.009 \\
 & 192 & 2.32M & 5/5 & 0.225$\pm$0.013 & 0.300$\pm$0.014 & 2.25M & 5/5 & 0.235$\pm$0.014 & 0.306$\pm$0.012 \\
 & 336 & 2.37M & 5/5 & 0.243$\pm$0.012 & 0.303$\pm$0.007 & 2.25M & 5/5 & 0.247$\pm$0.010 & 0.310$\pm$0.004 \\
 & 720 & 2.51M & 5/5 & 0.244$\pm$0.009 & 0.304$\pm$0.007 & 2.26M & 5/5 & 0.247$\pm$0.010 & 0.308$\pm$0.008 \\
\midrule
\multirow{4}{*}{Electricity} & 96 & 77.9K & 5/5 & 0.155$\pm$0.000 & 0.246$\pm$0.000 & 61.8K & 5/5 & 0.157$\pm$0.000 & 0.251$\pm$0.000 \\
 & 192 & 95.6K & 5/5 & 0.167$\pm$0.001 & 0.258$\pm$0.001 & 62.9K & 5/5 & 0.170$\pm$0.001 & 0.264$\pm$0.001 \\
 & 336 & 122.1K & 5/5 & 0.184$\pm$0.000 & 0.275$\pm$0.000 & 64.7K & 5/5 & 0.188$\pm$0.001 & 0.281$\pm$0.002 \\
 & 720 & 192.8K & 5/5 & 0.228$\pm$0.003 & 0.314$\pm$0.003 & 69.3K & 5/5 & 0.228$\pm$0.002 & 0.314$\pm$0.003 \\
\midrule
\multirow{4}{*}{Traffic} & 96 & 88.6K & 5/5 & 0.463$\pm$0.006 & 0.289$\pm$0.002 & 72.5K & 5/5 & 0.470$\pm$0.006 & 0.290$\pm$0.003 \\
 & 192 & 106.3K & 5/5 & 0.482$\pm$0.011 & 0.299$\pm$0.011 & 73.6K & 5/5 & 0.487$\pm$0.014 & 0.305$\pm$0.015 \\
 & 336 & 132.8K & 5/5 & 0.493$\pm$0.011 & 0.304$\pm$0.011 & 75.4K & 5/5 & 0.505$\pm$0.006 & 0.311$\pm$0.014 \\
 & 720 & 203.5K & 5/5 & 0.525$\pm$0.010 & 0.321$\pm$0.011 & 80.0K & 5/5 & 0.534$\pm$0.009 & 0.319$\pm$0.006 \\
\bottomrule
\end{tabular}
}
\caption{Paired TimeMixer and Lag-TimeMixer results over five seeds. TimeMixer follows the main-experiment configurations in its official paper; options not specified there follow the released implementation. Lag-TimeMixer uses the identical protocol without a separate search, preserves PDM and all components outside FMM, and replaces only the FMM temporal predictors with \laglayer{} operators. Params denotes total trainable parameters. Runs reports the number of completed seeds out of five.}
\label{tab:supp-timemixer-pair}
\end{table*}

\begin{table*}[t]
\centering
{\small
\setlength{\tabcolsep}{1mm}
\begin{tabular}{ll*{2}{cccc}}
\toprule
Dataset & $H$ & \multicolumn{4}{c}{TimeKAN} & \multicolumn{4}{c}{Lag-TimeKAN} \\
\cmidrule(lr){3-6}\cmidrule(lr){7-10}
& & Params & Runs & MSE & MAE & Params & Runs & MSE & MAE \\
\midrule
ETTh1 & 96 & 12.8K & 5/5 & 0.370$\pm$0.001 & 0.397$\pm$0.001 & 4.0K & 5/5 & 0.367$\pm$0.003 & 0.395$\pm$0.002 \\
 & 192 & 20.5K & 5/5 & 0.416$\pm$0.002 & 0.421$\pm$0.001 & 2.7K & 5/5 & 0.420$\pm$0.006 & 0.422$\pm$0.002 \\
 & 336 & 36.1K & 5/5 & 0.446$\pm$0.003 & 0.437$\pm$0.001 & 4.7K & 5/5 & 0.443$\pm$0.004 & 0.438$\pm$0.002 \\
 & 720 & 77.6K & 5/5 & 0.456$\pm$0.006 & 0.460$\pm$0.007 & 10.1K & 5/5 & 0.456$\pm$0.014 & 0.455$\pm$0.007 \\
\midrule
ETTh2 & 96 & 15.0K & 5/5 & 0.296$\pm$0.005 & 0.345$\pm$0.004 & 6.2K & 5/5 & 0.306$\pm$0.005 & 0.356$\pm$0.004 \\
 & 192 & 19.6K & 5/5 & 0.380$\pm$0.005 & 0.396$\pm$0.004 & 1.8K & 5/5 & 0.383$\pm$0.005 & 0.397$\pm$0.004 \\
 & 336 & 43.5K & 5/5 & 0.396$\pm$0.006 & 0.419$\pm$0.004 & 12.1K & 5/5 & 0.410$\pm$0.005 & 0.430$\pm$0.003 \\
 & 720 & 77.6K & 5/5 & 0.431$\pm$0.008 & 0.444$\pm$0.005 & 10.1K & 5/5 & 0.444$\pm$0.012 & 0.454$\pm$0.009 \\
\midrule
ETTm1 & 96 & 14.4K & 5/5 & 0.330$\pm$0.003 & 0.367$\pm$0.002 & 5.5K & 5/5 & 0.333$\pm$0.008 & 0.370$\pm$0.005 \\
 & 192 & 23.7K & 5/5 & 0.359$\pm$0.003 & 0.384$\pm$0.002 & 5.8K & 5/5 & 0.370$\pm$0.008 & 0.390$\pm$0.004 \\
 & 336 & 40.1K & 5/5 & 0.388$\pm$0.003 & 0.405$\pm$0.003 & 8.7K & 5/5 & 0.390$\pm$0.004 & 0.405$\pm$0.002 \\
 & 720 & 77.6K & 5/5 & 0.448$\pm$0.003 & 0.439$\pm$0.002 & 10.1K & 5/5 & 0.455$\pm$0.006 & 0.441$\pm$0.004 \\
\midrule
ETTm2 & 96 & 38.1K & 5/5 & 0.175$\pm$0.001 & 0.257$\pm$0.001 & 29.3K & 5/5 & 0.182$\pm$0.003 & 0.268$\pm$0.004 \\
 & 192 & 47.4K & 5/5 & 0.243$\pm$0.003 & 0.305$\pm$0.002 & 29.6K & 5/5 & 0.255$\pm$0.004 & 0.317$\pm$0.003 \\
 & 336 & 38.2K & 5/5 & 0.305$\pm$0.004 & 0.344$\pm$0.002 & 6.8K & 5/5 & 0.316$\pm$0.007 & 0.355$\pm$0.006 \\
 & 720 & 81.4K & 5/5 & 0.413$\pm$0.010 & 0.411$\pm$0.009 & 13.9K & 5/5 & 0.415$\pm$0.010 & 0.415$\pm$0.007 \\
\midrule
Weather & 96 & 20.9K & 5/5 & 0.164$\pm$0.001 & 0.211$\pm$0.001 & 12.1K & 5/5 & 0.167$\pm$0.002 & 0.216$\pm$0.002 \\
 & 192 & 26.2K & 5/5 & 0.208$\pm$0.001 & 0.251$\pm$0.000 & 8.4K & 5/5 & 0.214$\pm$0.001 & 0.260$\pm$0.003 \\
 & 336 & 40.2K & 5/5 & 0.263$\pm$0.001 & 0.291$\pm$0.001 & 8.8K & 5/5 & 0.268$\pm$0.001 & 0.298$\pm$0.003 \\
 & 720 & 77.4K & 5/5 & 0.339$\pm$0.001 & 0.343$\pm$0.002 & 10.0K & 5/5 & 0.342$\pm$0.002 & 0.347$\pm$0.002 \\
\midrule
Solar & 96 & 21.9K & 5/5 & 0.239$\pm$0.008 & 0.287$\pm$0.003 & 13.0K & 5/5 & 0.252$\pm$0.017 & 0.305$\pm$0.015 \\
 & 192 & 31.2K & 5/5 & 0.279$\pm$0.009 & 0.316$\pm$0.004 & 13.3K & 5/5 & 0.305$\pm$0.019 & 0.334$\pm$0.010 \\
 & 336 & 45.1K & 5/5 & 0.310$\pm$0.019 & 0.334$\pm$0.012 & 13.8K & 5/5 & 0.342$\pm$0.016 & 0.355$\pm$0.006 \\
 & 720 & 82.4K & 5/5 & 0.300$\pm$0.002 & 0.322$\pm$0.003 & 14.9K & 5/5 & 0.346$\pm$0.024 & 0.361$\pm$0.007 \\
\midrule
Electricity & 96 & 23.3K & 5/5 & 0.176$\pm$0.001 & 0.269$\pm$0.002 & 14.5K & 5/5 & 0.182$\pm$0.002 & 0.274$\pm$0.002 \\
 & 192 & 32.6K & 5/5 & 0.182$\pm$0.001 & 0.273$\pm$0.001 & 14.8K & 5/5 & 0.187$\pm$0.001 & 0.277$\pm$0.001 \\
 & 336 & 46.6K & 5/5 & 0.198$\pm$0.001 & 0.289$\pm$0.001 & 15.2K & 5/5 & 0.201$\pm$0.001 & 0.291$\pm$0.001 \\
 & 720 & 83.9K & 5/5 & 0.241$\pm$0.002 & 0.324$\pm$0.002 & 16.4K & 5/5 & 0.244$\pm$0.002 & 0.325$\pm$0.002 \\
\midrule
Traffic & 96 & 27.7K & 5/5 & 0.585$\pm$0.012 & 0.380$\pm$0.003 & 18.8K & 5/5 & 0.591$\pm$0.006 & 0.378$\pm$0.008 \\
 & 192 & 37.0K & 5/5 & 0.559$\pm$0.006 & 0.363$\pm$0.001 & 19.1K & 5/5 & 0.579$\pm$0.006 & 0.364$\pm$0.007 \\
 & 336 & 50.9K & 5/5 & 0.571$\pm$0.005 & 0.364$\pm$0.002 & 19.6K & 5/5 & 0.589$\pm$0.006 & 0.366$\pm$0.004 \\
 & 720 & 88.2K & 5/5 & 0.603$\pm$0.007 & 0.381$\pm$0.002 & 20.7K & 5/5 & 0.622$\pm$0.018 & 0.390$\pm$0.004 \\
\bottomrule
\end{tabular}
}
\caption{Paired TimeKAN and Lag-TimeKAN results under the unified training protocol. Each entry reports mean$\pm$standard deviation and total trainable parameters. Lag-TimeKAN uses the identical data, optimization, and random-seed setting as TimeKAN and replaces only the final linear temporal predictor with the complete \laglayer{} operator. Runs reports the number of completed seeds out of five.}
\label{tab:supp-timekan-lag-results}
\end{table*}

\subsection{Standalone Lag-Structured Results}

Table~\ref{tab:supp-ours-full-results} gives the complete \laglayer{} and \method{} batch results.
Both columns use the final selected lag-structured runs with grouped lag regularization; their dataset-specific batch sizes are listed in Table~\ref{tab:supp-main-settings}.

\begin{table*}[t]
\centering
{\small
\setlength{\tabcolsep}{1mm}
\begin{tabular}{ll*{8}{c}}
\toprule
Dataset & $H$ & \multicolumn{4}{c}{\laglayer{}} & \multicolumn{4}{c}{\method{}} \\
\cmidrule(lr){3-6}\cmidrule(lr){7-10}
& & Params & Runs & MSE & MAE & Params & Runs & MSE & MAE \\
\midrule
ETTh1 & 96 & 479 & 5/5 & 0.390$\pm$0.001 & 0.396$\pm$0.001 & 3.5K & 5/5 & 0.389$\pm$0.002 & 0.402$\pm$0.002 \\
 & 192 & 767 & 5/5 & 0.440$\pm$0.001 & 0.425$\pm$0.000 & 6.3K & 5/5 & 0.438$\pm$0.004 & 0.430$\pm$0.002 \\
 & 336 & 1.2K & 5/5 & 0.483$\pm$0.001 & 0.447$\pm$0.000 & 10.5K & 5/5 & 0.487$\pm$0.005 & 0.448$\pm$0.002 \\
 & 720 & 2.4K & 5/5 & 0.482$\pm$0.000 & 0.470$\pm$0.000 & 21.6K & 5/5 & 0.531$\pm$0.020 & 0.493$\pm$0.009 \\
\midrule
ETTh2 & 96 & 479 & 5/5 & 0.295$\pm$0.002 & 0.341$\pm$0.001 & 3.5K & 5/5 & 0.301$\pm$0.009 & 0.348$\pm$0.006 \\
 & 192 & 767 & 5/5 & 0.378$\pm$0.001 & 0.391$\pm$0.001 & 6.3K & 5/5 & 0.369$\pm$0.004 & 0.391$\pm$0.004 \\
 & 336 & 1.2K & 5/5 & 0.417$\pm$0.001 & 0.426$\pm$0.000 & 10.5K & 5/5 & 0.419$\pm$0.004 & 0.432$\pm$0.002 \\
 & 720 & 2.4K & 5/5 & 0.421$\pm$0.000 & 0.439$\pm$0.000 & 21.6K & 5/5 & 0.447$\pm$0.008 & 0.460$\pm$0.005 \\
\midrule
ETTm1 & 96 & 3.4K & 5/5 & 0.352$\pm$0.001 & 0.374$\pm$0.001 & 24.8K & 5/5 & 0.324$\pm$0.002 & 0.363$\pm$0.001 \\
 & 192 & 5.4K & 5/5 & 0.389$\pm$0.001 & 0.390$\pm$0.000 & 44.3K & 5/5 & 0.369$\pm$0.004 & 0.389$\pm$0.003 \\
 & 336 & 8.4K & 5/5 & 0.418$\pm$0.000 & 0.409$\pm$0.000 & 73.5K & 5/5 & 0.397$\pm$0.005 & 0.409$\pm$0.004 \\
 & 720 & 16.5K & 5/5 & 0.481$\pm$0.001 & 0.442$\pm$0.001 & 151.5K & 5/5 & 0.461$\pm$0.005 & 0.446$\pm$0.005 \\
\midrule
ETTm2 & 96 & 3.4K & 5/5 & 0.179$\pm$0.001 & 0.261$\pm$0.001 & 24.8K & 5/5 & 0.178$\pm$0.001 & 0.259$\pm$0.001 \\
 & 192 & 5.4K & 5/5 & 0.245$\pm$0.001 & 0.302$\pm$0.001 & 44.3K & 5/5 & 0.242$\pm$0.001 & 0.301$\pm$0.001 \\
 & 336 & 8.4K & 5/5 & 0.307$\pm$0.001 & 0.342$\pm$0.000 & 73.5K & 5/5 & 0.300$\pm$0.001 & 0.340$\pm$0.002 \\
 & 720 & 16.5K & 5/5 & 0.409$\pm$0.001 & 0.398$\pm$0.000 & 151.5K & 5/5 & 0.403$\pm$0.003 & 0.400$\pm$0.002 \\
\midrule
Weather & 96 & 10.1K & 5/5 & 0.164$\pm$0.001 & 0.208$\pm$0.001 & 74.4K & 5/5 & 0.165$\pm$0.001 & 0.213$\pm$0.001 \\
 & 192 & 16.1K & 5/5 & 0.210$\pm$0.001 & 0.249$\pm$0.000 & 132.9K & 5/5 & 0.212$\pm$0.002 & 0.256$\pm$0.002 \\
 & 336 & 25.2K & 5/5 & 0.265$\pm$0.001 & 0.290$\pm$0.001 & 220.6K & 5/5 & 0.269$\pm$0.002 & 0.296$\pm$0.002 \\
 & 720 & 49.4K & 5/5 & 0.344$\pm$0.001 & 0.340$\pm$0.001 & 454.4K & 5/5 & 0.350$\pm$0.004 & 0.348$\pm$0.003 \\
\midrule
Solar & 96 & 479 & 5/5 & 0.321$\pm$0.002 & 0.341$\pm$0.002 & 123.0K & 5/5 & 0.223$\pm$0.005 & 0.263$\pm$0.004 \\
 & 192 & 767 & 5/5 & 0.358$\pm$0.002 & 0.360$\pm$0.002 & 244.1K & 5/5 & 0.253$\pm$0.017 & 0.279$\pm$0.010 \\
 & 336 & 1.2K & 5/5 & 0.397$\pm$0.001 & 0.371$\pm$0.001 & 425.7K & 5/5 & 0.273$\pm$0.004 & 0.294$\pm$0.005 \\
 & 720 & 2.4K & 5/5 & 0.397$\pm$0.001 & 0.362$\pm$0.001 & 909.9K & 5/5 & 0.270$\pm$0.004 & 0.291$\pm$0.002 \\
\midrule
Electricity & 96 & 479 & 5/5 & 0.203$\pm$0.000 & 0.279$\pm$0.001 & 123.0K & 5/5 & 0.167$\pm$0.001 & 0.255$\pm$0.001 \\
 & 192 & 767 & 5/5 & 0.201$\pm$0.000 & 0.280$\pm$0.001 & 244.1K & 5/5 & 0.174$\pm$0.001 & 0.262$\pm$0.001 \\
 & 336 & 1.2K & 5/5 & 0.215$\pm$0.000 & 0.294$\pm$0.000 & 425.7K & 5/5 & 0.189$\pm$0.000 & 0.278$\pm$0.001 \\
 & 720 & 2.4K & 5/5 & 0.257$\pm$0.000 & 0.327$\pm$0.001 & 909.9K & 5/5 & 0.232$\pm$0.002 & 0.316$\pm$0.001 \\
\midrule
Traffic & 96 & 479 & 5/5 & 0.653$\pm$0.000 & 0.386$\pm$0.001 & 123.0K & 5/5 & 0.439$\pm$0.003 & 0.287$\pm$0.003 \\
 & 192 & 767 & 5/5 & 0.607$\pm$0.001 & 0.366$\pm$0.001 & 244.1K & 5/5 & 0.444$\pm$0.002 & 0.284$\pm$0.002 \\
 & 336 & 1.2K & 5/5 & 0.616$\pm$0.000 & 0.369$\pm$0.001 & 425.7K & 5/5 & 0.456$\pm$0.001 & 0.292$\pm$0.002 \\
 & 720 & 2.4K & 5/5 & 0.653$\pm$0.000 & 0.389$\pm$0.001 & 909.9K & 5/5 & 0.491$\pm$0.004 & 0.309$\pm$0.002 \\
\bottomrule
\end{tabular}
}
\caption{Complete \laglayer{} and \method{} main results. Metrics are reported as mean$\pm$standard deviation over five seeds. Params denotes the total number of trainable parameters in each instantiated model. \laglayer{} is the complete single-layer operator with both horizon-wise and history-wise scaling, while \method{} is its deep nonlinear extension.}
\label{tab:supp-ours-full-results}
\end{table*}

\section{Complete Ablation Results}
\label{app:complete-ablation}

This section provides the complete ablation results corresponding to the compact ablation table in the main paper.
The single-layer variants first isolate relative-lag sharing and each scale
mechanism. The deep entry is a separate capacity diagnostic in which both
scale vectors are disabled, so that its difference from the unscaled
single-layer backbone isolates nonlinear composition and multiple feature
maps. This diagnostic setting should be distinguished from the complete
\method{} used in the main forecasting results, where every layer includes
adaptive scaling.
All variants in this diagnostic ablation use the fixed diagnostic seed 2027.
All entries use a shared channel-independent operator, batch size 64, Adam
with learning rate $10^{-3}$, at most 20 epochs, early-stopping patience 5,
and the preprocessing and evaluation protocol in
Section~\ref{app:experimental-settings}.
The main paper reports dataset-wise averages for compactness, while
Table~\ref{tab:supp-ablation-full} reports the full horizon-wise MSE and MAE.
The purpose of this ablation is not to tune a different model for each dataset, but to show how each structural component changes the behavior of the same lag-structured forecasting family.

\begin{table*}[t]
\centering
{\small
\setlength{\tabcolsep}{1mm}
\begin{tabular}{lcccccc}
\toprule
Variant & \shortstack{Relative-lag\\sharing} & \shortstack{Row\\scale} &
\shortstack{Column\\scale} & \shortstack{Nonlinear\\depth} &
\shortstack{Extra\\modules} & Diagnostic \\
\midrule
Linear & No & No & No & No & No & free matrix \\
\laglayer{} w/o Scaling & Yes & No & No & No & No & lag sharing \\
\laglayer{} w/o History Scale & Yes & Yes & No & No & No & row adaptation \\
\laglayer{} w/o Horizon Scale & Yes & No & Yes & No & No & column adaptation \\
\laglayer{} & Yes & Yes & Yes & No & No & row/column adaptation \\
\method{} capacity diagnostic & Yes & No & No & Yes & No & nonlinear responses \\
\bottomrule
\end{tabular}
}
\caption{Ablation variants and the component isolated by each diagnostic. The deep capacity diagnostic disables adaptive scaling; the complete \method{} used in the main forecasting table enables it in every layer. No extra decomposition, attention, gating, or residual branch is introduced.}
\label{tab:supp-ablation-components}
\end{table*}

\begin{table*}[t]
\centering
{\small
\setlength{\tabcolsep}{0.75mm}
\begin{tabular}{llcccccc}
\toprule
Dataset & $H$ & Linear & \shortstack{w/o\\Scaling} & \shortstack{w/o History\\Scale} & \shortstack{w/o Horizon\\Scale} & \laglayer{} & \method{} \\
\midrule
\multirow{4}{*}{ETTh1} & 96 & 0.387/0.394 & 0.393/0.397 & 0.391/0.395 & 0.393/0.398 & 0.392/0.397 & 0.389/0.401 \\
 & 192 & 0.438/0.424 & 0.439/0.424 & 0.438/0.423 & 0.440/0.425 & 0.440/0.424 & 0.441/0.434 \\
 & 336 & 0.482/0.447 & 0.482/0.446 & 0.482/0.445 & 0.482/0.447 & 0.482/0.447 & 0.491/0.456 \\
 & 720 & 0.482/0.471 & 0.480/0.469 & 0.480/0.469 & 0.481/0.471 & 0.482/0.470 & 0.498/0.477 \\
\midrule
\multirow{4}{*}{ETTh2} & 96 & 0.290/0.338 & 0.304/0.345 & 0.303/0.344 & 0.294/0.341 & 0.294/0.340 & 0.301/0.345 \\
 & 192 & 0.376/0.391 & 0.389/0.396 & 0.388/0.395 & 0.378/0.391 & 0.377/0.391 & 0.369/0.390 \\
 & 336 & 0.418/0.427 & 0.425/0.430 & 0.425/0.429 & 0.416/0.426 & 0.416/0.426 & 0.409/0.427 \\
 & 720 & 0.422/0.440 & 0.427/0.441 & 0.427/0.441 & 0.421/0.440 & 0.421/0.439 & 0.446/0.455 \\
\midrule
\multirow{4}{*}{ETTm1} & 96 & 0.352/0.372 & 0.377/0.390 & 0.375/0.388 & 0.376/0.389 & 0.371/0.383 & 0.341/0.373 \\
 & 192 & 0.391/0.391 & 0.407/0.400 & 0.404/0.398 & 0.405/0.401 & 0.402/0.397 & 0.381/0.392 \\
 & 336 & 0.423/0.413 & 0.436/0.420 & 0.435/0.419 & 0.436/0.420 & 0.434/0.418 & 0.411/0.415 \\
 & 720 & 0.487/0.448 & 0.496/0.452 & 0.495/0.453 & 0.496/0.453 & 0.495/0.453 & 0.476/0.447 \\
\midrule
\multirow{4}{*}{ETTm2} & 96 & 0.181/0.263 & 0.193/0.273 & 0.190/0.269 & 0.183/0.267 & 0.187/0.267 & 0.182/0.265 \\
 & 192 & 0.246/0.304 & 0.256/0.311 & 0.253/0.308 & 0.249/0.307 & 0.248/0.306 & 0.246/0.307 \\
 & 336 & 0.307/0.342 & 0.316/0.347 & 0.314/0.346 & 0.311/0.347 & 0.311/0.346 & 0.304/0.344 \\
 & 720 & 0.407/0.398 & 0.414/0.401 & 0.413/0.400 & 0.412/0.401 & 0.411/0.401 & 0.417/0.410 \\
\midrule
\multirow{4}{*}{Weather} & 96 & 0.193/0.233 & 0.208/0.251 & 0.202/0.241 & 0.193/0.237 & 0.195/0.235 & 0.190/0.230 \\
 & 192 & 0.238/0.269 & 0.252/0.284 & 0.247/0.275 & 0.244/0.278 & 0.242/0.272 & 0.236/0.270 \\
 & 336 & 0.291/0.307 & 0.302/0.317 & 0.298/0.311 & 0.294/0.312 & 0.291/0.307 & 0.285/0.302 \\
 & 720 & 0.365/0.354 & 0.372/0.361 & 0.369/0.356 & 0.366/0.357 & 0.364/0.353 & 0.357/0.350 \\
\midrule
\multirow{4}{*}{Solar} & 96 & 0.307/0.330 & 0.522/0.454 & 0.408/0.369 & 0.378/0.425 & 0.325/0.348 & 0.218/0.263 \\
 & 192 & 0.347/0.351 & 0.549/0.467 & 0.436/0.378 & 0.468/0.477 & 0.360/0.363 & 0.255/0.282 \\
 & 336 & 0.387/0.365 & 0.561/0.471 & 0.463/0.390 & 0.497/0.482 & 0.399/0.376 & 0.287/0.300 \\
 & 720 & 0.389/0.359 & 0.534/0.454 & 0.452/0.381 & 0.488/0.467 & 0.400/0.368 & 0.262/0.285 \\
\midrule
\multirow{4}{*}{Electricity} & 96 & 0.198/0.275 & 0.215/0.290 & 0.209/0.280 & 0.208/0.286 & 0.203/0.279 & 0.168/0.255 \\
 & 192 & 0.198/0.278 & 0.211/0.290 & 0.205/0.282 & 0.207/0.287 & 0.201/0.280 & 0.173/0.261 \\
 & 336 & 0.212/0.292 & 0.224/0.304 & 0.219/0.296 & 0.221/0.302 & 0.215/0.295 & 0.190/0.278 \\
 & 720 & 0.254/0.325 & 0.267/0.336 & 0.261/0.329 & 0.262/0.333 & 0.257/0.328 & 0.234/0.315 \\
\midrule
\multirow{4}{*}{Traffic} & 96 & 0.646/0.385 & 0.705/0.425 & 0.668/0.395 & 0.653/0.387 & 0.653/0.388 & 0.439/0.288 \\
 & 192 & 0.599/0.362 & 0.654/0.405 & 0.614/0.369 & 0.627/0.384 & 0.608/0.367 & 0.444/0.284 \\
 & 336 & 0.606/0.364 & 0.661/0.405 & 0.623/0.370 & 0.643/0.396 & 0.620/0.369 & 0.458/0.291 \\
 & 720 & 0.644/0.384 & 0.702/0.424 & 0.660/0.390 & 0.684/0.417 & 0.654/0.389 & 0.498/0.313 \\
\bottomrule
\end{tabular}
}
\caption{Complete horizon-wise MSE/MAE component diagnostics. Each entry reports MSE/MAE. Linear is the unconstrained free-matrix reference. For \laglayer{}, ``w/o Scaling'' removes both scale vectors, while each one-sided ablation removes the named scale. The \method{} capacity diagnostic adds nonlinear lag-structured composition to the unscaled backbone.}
\label{tab:supp-ablation-full}
\end{table*}

\section{Depth and Width Analysis}
\label{app:depth-width-analysis}

This section reports the full depth/width sweep for the deep \method{} architecture on Solar, Electricity, and Traffic.
The sweep and its associated qualitative forecasting case use the fixed diagnostic seed 2027.
The depth sweep fixes the hidden width to $M=8$ and varies the number of lag-structured layers.
The width sweep fixes the depth to $R=3$ and varies the hidden width, where $M$ denotes the number of hidden feature maps and $R$ denotes the number of lag-structured layers.
Both scale mechanisms are disabled throughout this sweep, while batch size 64
and the remaining standalone optimization settings are fixed. This protocol
isolates whether the lag-structured-layer--activation--lag-structured-layer
construction provides useful nonlinear capacity rather than merely increasing
parameters.
Table~\ref{tab:supp-depth-width-full} reports the horizon-wise MSE and MAE.

\begin{table*}[t]
\centering
\textit{(a) MSE}\\
{\small
\setlength{\tabcolsep}{1mm}
\begin{tabular}{llccccccccc}
\toprule
Dataset & $H$ & R2-M8 & R3-M8 & R4-M8 & R5-M8 & R3-M1 & R3-M2 & R3-M4 & R3-M16 & R3-M32 \\
\midrule
\multirow{4}{*}{Solar} & 96 & 0.256 & 0.230 & 0.224 & 0.226 & 0.287 & 0.245 & 0.240 & 0.227 & 0.213 \\
 & 192 & 0.289 & 0.268 & 0.263 & 0.261 & 0.318 & 0.295 & 0.278 & 0.260 & 0.257 \\
 & 336 & 0.331 & 0.280 & 0.273 & 0.284 & 0.345 & 0.302 & 0.306 & 0.284 & 0.289 \\
 & 720 & 0.317 & 0.285 & 0.265 & 0.269 & 0.322 & 0.306 & 0.293 & 0.268 & 0.310 \\
\midrule
\multirow{4}{*}{Electricity} & 96 & 0.188 & 0.180 & 0.176 & 0.174 & 0.206 & 0.196 & 0.187 & 0.172 & 0.169 \\
 & 192 & 0.192 & 0.186 & 0.178 & 0.179 & 0.206 & 0.195 & 0.188 & 0.176 & 0.173 \\
 & 336 & 0.205 & 0.197 & 0.193 & 0.191 & 0.218 & 0.207 & 0.202 & 0.193 & 0.194 \\
 & 720 & 0.246 & 0.240 & 0.231 & 0.234 & 0.254 & 0.248 & 0.241 & 0.233 & 0.233 \\
\midrule
\multirow{4}{*}{Traffic} & 96 & 0.527 & 0.475 & 0.460 & 0.459 & 0.633 & 0.543 & 0.492 & 0.456 & 0.440 \\
 & 192 & 0.505 & 0.470 & 0.458 & 0.452 & 0.577 & 0.517 & 0.486 & 0.455 & 0.448 \\
 & 336 & 0.514 & 0.476 & 0.469 & 0.459 & 0.546 & 0.505 & 0.488 & 0.468 & 0.469 \\
 & 720 & 0.544 & 0.504 & 0.493 & 0.487 & 0.549 & 0.522 & 0.507 & 0.500 & 0.509 \\
\bottomrule
\end{tabular}
}
\par\medskip
\textit{(b) MAE}\\
\centering
{\small
\setlength{\tabcolsep}{1mm}
\begin{tabular}{llccccccccc}
\toprule
Dataset & $H$ & R2-M8 & R3-M8 & R4-M8 & R5-M8 & R3-M1 & R3-M2 & R3-M4 & R3-M16 & R3-M32 \\
\midrule
\multirow{4}{*}{Solar} & 96 & 0.296 & 0.275 & 0.265 & 0.267 & 0.327 & 0.287 & 0.280 & 0.275 & 0.257 \\
 & 192 & 0.315 & 0.294 & 0.293 & 0.280 & 0.342 & 0.325 & 0.307 & 0.291 & 0.285 \\
 & 336 & 0.338 & 0.299 & 0.296 & 0.299 & 0.360 & 0.317 & 0.319 & 0.308 & 0.310 \\
 & 720 & 0.328 & 0.304 & 0.288 & 0.284 & 0.343 & 0.318 & 0.312 & 0.298 & 0.322 \\
\midrule
\multirow{4}{*}{Electricity} & 96 & 0.271 & 0.267 & 0.262 & 0.260 & 0.285 & 0.280 & 0.270 & 0.259 & 0.255 \\
 & 192 & 0.275 & 0.272 & 0.265 & 0.266 & 0.288 & 0.278 & 0.272 & 0.264 & 0.260 \\
 & 336 & 0.289 & 0.286 & 0.281 & 0.279 & 0.300 & 0.291 & 0.288 & 0.279 & 0.281 \\
 & 720 & 0.322 & 0.318 & 0.317 & 0.316 & 0.328 & 0.324 & 0.319 & 0.314 & 0.313 \\
\midrule
\multirow{4}{*}{Traffic} & 96 & 0.344 & 0.313 & 0.300 & 0.303 & 0.426 & 0.365 & 0.325 & 0.301 & 0.286 \\
 & 192 & 0.328 & 0.301 & 0.294 & 0.289 & 0.388 & 0.339 & 0.318 & 0.293 & 0.284 \\
 & 336 & 0.331 & 0.299 & 0.299 & 0.294 & 0.363 & 0.326 & 0.312 & 0.298 & 0.298 \\
 & 720 & 0.347 & 0.322 & 0.311 & 0.310 & 0.353 & 0.335 & 0.325 & 0.317 & 0.320 \\
\bottomrule
\end{tabular}
}
\caption{Complete horizon-wise depth and width sweeps for the unscaled capacity diagnostic. R$r$-M$m$ denotes $r$ lag-structured layers and $m$ hidden feature maps.}
\label{tab:supp-depth-width-full}
\end{table*}

\begin{figure}[t]
\centering
\includegraphics[width=0.98\columnwidth,draft=false]{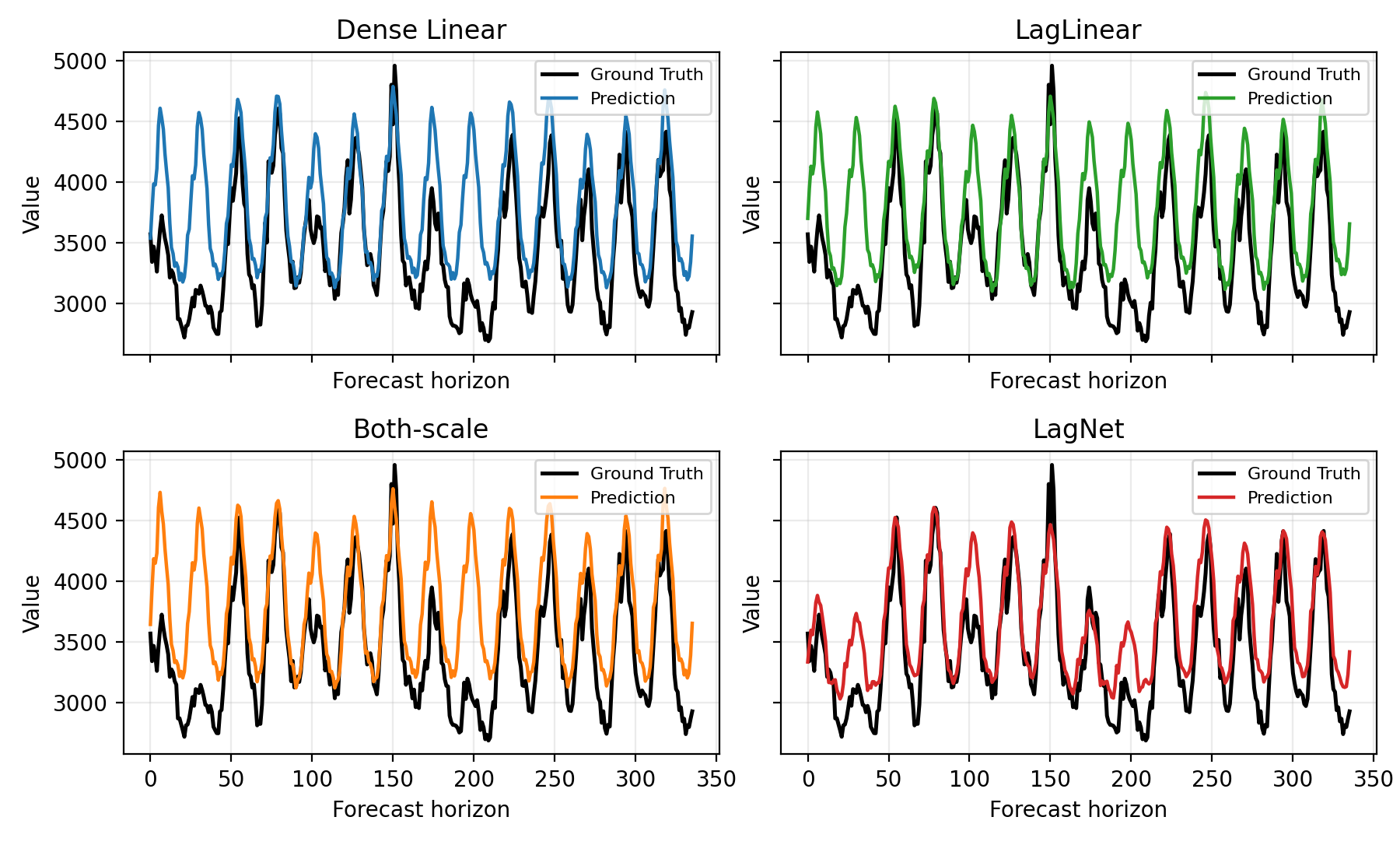}
\caption{Forecasts on a fixed Electricity test case with $H=336$, used as a qualitative complement to the unscaled depth/width diagnostic.}
\label{fig:prediction-case}
\end{figure}

Figure~\ref{fig:prediction-case} provides a qualitative complement to the
capacity sweeps using a fixed Electricity test window with $H=336$.
In this case, the unscaled deep configuration follows changes in level and
amplitude more closely than Linear and the single-layer lag-structured
variants, illustrating the additional fitting capacity provided by depth and
hidden feature maps.

\section{Sample-Efficiency Protocol and Results}
\label{app:sample-efficiency}
The sample-efficiency experiment trains representative models using fractions of the original training windows.
Validation and test splits remain unchanged, and only the training windows are subsampled.
We sample without replacement using a deterministic permutation for each
fraction and fixed seed 2027; no minimum subset size is imposed.
The training fractions are 1\%, 5\%, 20\%, 50\%, and 100\%.
All models use look-back length $L=96$, prediction lengths $H\in\{96,192,336,720\}$, RevIN normalization, scaled metric space, batch size 64, Adam optimization with learning rate $10^{-3}$, at most 20 epochs, early-stopping patience 5, gradient clipping at 1.0, and automatic mixed precision.
The compared models are Linear, \laglayer{}, and \method{}.
Linear is included as an unconstrained free-matrix reference; \laglayer{} tests the complete single-layer lag-structured operator with horizon-wise and history-wise scaling; \method{} tests its nonlinear deep extension.
For \method{}, the five small datasets use two layers and four hidden feature maps, while Solar, Electricity, and Traffic use four layers and 12 hidden feature maps, matching the corresponding main-experiment architectures.
The global and lag-kernel weight decays are $10^{-5}$, while scale coefficients and biases are not decayed.

Table~\ref{tab:supp-sample-efficiency-mse} reports the complete horizon-wise MSE results for all eight datasets.

\begin{table*}[p]
\centering
{\small
\setlength{\tabcolsep}{1mm}
\begin{tabular}{@{}c@{\hspace{2mm}}c@{}}
\begin{tabular}[t]{ll l ccccc}
\toprule
Dataset & $H$ & Model & 1\% & 5\% & 20\% & 50\% & 100\% \\
\midrule
ETTh1 & 96 & Linear & 0.490 & 0.419 & 0.394 & 0.391 & 0.386 \\
 &  & LagLayer & 0.421 & 0.399 & 0.391 & 0.394 & 0.391 \\
 &  & LagNet & 0.429 & 0.404 & 0.391 & 0.385 & 0.388 \\
\addlinespace[0.2em]
 & 192 & Linear & 0.546 & 0.468 & 0.446 & 0.440 & 0.441 \\
 &  & LagLayer & 0.474 & 0.447 & 0.441 & 0.441 & 0.441 \\
 &  & LagNet & 0.455 & 0.444 & 0.442 & 0.436 & 0.432 \\
\addlinespace[0.2em]
 & 336 & Linear & 0.579 & 0.509 & 0.487 & 0.483 & 0.482 \\
 &  & LagLayer & 0.508 & 0.486 & 0.481 & 0.485 & 0.486 \\
 &  & LagNet & 0.516 & 0.481 & 0.502 & 0.489 & 0.498 \\
\addlinespace[0.2em]
 & 720 & Linear & 0.571 & 0.513 & 0.486 & 0.486 & 0.482 \\
 &  & LagLayer & 0.489 & 0.479 & 0.479 & 0.485 & 0.482 \\
 &  & LagNet & 0.496 & 0.480 & 0.569 & 0.569 & 0.567 \\
\midrule
ETTh2 & 96 & Linear & 0.362 & 0.313 & 0.297 & 0.292 & 0.292 \\
 &  & LagLayer & 0.321 & 0.307 & 0.302 & 0.296 & 0.292 \\
 &  & LagNet & 0.350 & 0.299 & 0.295 & 0.296 & 0.314 \\
\addlinespace[0.2em]
 & 192 & Linear & 0.439 & 0.401 & 0.387 & 0.378 & 0.375 \\
 &  & LagLayer & 0.403 & 0.391 & 0.384 & 0.378 & 0.377 \\
 &  & LagNet & 0.401 & 0.376 & 0.373 & 0.363 & 0.364 \\
\addlinespace[0.2em]
 & 336 & Linear & 0.463 & 0.436 & 0.426 & 0.423 & 0.416 \\
 &  & LagLayer & 0.430 & 0.423 & 0.420 & 0.416 & 0.416 \\
 &  & LagNet & 0.427 & 0.419 & 0.415 & 0.424 & 0.424 \\
\addlinespace[0.2em]
 & 720 & Linear & 0.471 & 0.458 & 0.428 & 0.427 & 0.423 \\
 &  & LagLayer & 0.428 & 0.432 & 0.420 & 0.423 & 0.421 \\
 &  & LagNet & 0.452 & 0.478 & 0.458 & 0.457 & 0.441 \\
\midrule
Weather & 96 & Linear & 0.212 & 0.198 & 0.195 & 0.192 & 0.194 \\
 &  & LagLayer & 0.211 & 0.201 & 0.195 & 0.196 & 0.194 \\
 &  & LagNet & 0.205 & 0.199 & 0.192 & 0.184 & 0.184 \\
\addlinespace[0.2em]
 & 192 & Linear & 0.254 & 0.243 & 0.239 & 0.241 & 0.239 \\
 &  & LagLayer & 0.253 & 0.247 & 0.243 & 0.240 & 0.240 \\
 &  & LagNet & 0.246 & 0.243 & 0.241 & 0.235 & 0.231 \\
\addlinespace[0.2em]
 & 336 & Linear & 0.306 & 0.294 & 0.292 & 0.291 & 0.291 \\
 &  & LagLayer & 0.303 & 0.295 & 0.292 & 0.292 & 0.291 \\
 &  & LagNet & 0.300 & 0.295 & 0.288 & 0.285 & 0.286 \\
\addlinespace[0.2em]
 & 720 & Linear & 0.377 & 0.367 & 0.364 & 0.365 & 0.365 \\
 &  & LagLayer & 0.373 & 0.366 & 0.364 & 0.363 & 0.364 \\
 &  & LagNet & 0.371 & 0.367 & 0.360 & 0.357 & 0.356 \\
\midrule
Solar & 96 & Linear & 0.393 & 0.336 & 0.312 & 0.308 & 0.305 \\
 &  & LagLayer & 0.508 & 0.396 & 0.344 & 0.327 & 0.322 \\
 &  & LagNet & 0.273 & 0.254 & 0.241 & 0.225 & 0.214 \\
\addlinespace[0.2em]
 & 192 & Linear & 0.426 & 0.373 & 0.352 & 0.345 & 0.343 \\
 &  & LagLayer & 0.546 & 0.435 & 0.382 & 0.365 & 0.357 \\
 &  & LagNet & 0.314 & 0.298 & 0.259 & 0.257 & 0.242 \\
\addlinespace[0.2em]
 & 336 & Linear & 0.453 & 0.409 & 0.393 & 0.389 & 0.386 \\
 &  & LagLayer & 0.561 & 0.468 & 0.416 & 0.401 & 0.395 \\
 &  & LagNet & 0.324 & 0.294 & 0.286 & 0.295 & 0.260 \\
\addlinespace[0.2em]
 & 720 & Linear & 0.452 & 0.408 & 0.394 & 0.391 & 0.388 \\
 &  & LagLayer & 0.545 & 0.476 & 0.413 & 0.401 & 0.399 \\
 &  & LagNet & 0.324 & 0.310 & 0.291 & 0.267 & 0.269 \\
\bottomrule
\end{tabular}
&
\begin{tabular}[t]{ll l ccccc}
\toprule
Dataset & $H$ & Model & 1\% & 5\% & 20\% & 50\% & 100\% \\
\midrule
ETTm1 & 96 & Linear & 0.395 & 0.364 & 0.352 & 0.349 & 0.356 \\
 &  & LagLayer & 0.398 & 0.380 & 0.366 & 0.368 & 0.368 \\
 &  & LagNet & 0.376 & 0.345 & 0.344 & 0.335 & 0.342 \\
\addlinespace[0.2em]
 & 192 & Linear & 0.421 & 0.396 & 0.392 & 0.393 & 0.391 \\
 &  & LagLayer & 0.419 & 0.411 & 0.405 & 0.402 & 0.404 \\
 &  & LagNet & 0.397 & 0.385 & 0.394 & 0.375 & 0.380 \\
\addlinespace[0.2em]
 & 336 & Linear & 0.452 & 0.432 & 0.428 & 0.424 & 0.423 \\
 &  & LagLayer & 0.442 & 0.435 & 0.434 & 0.433 & 0.434 \\
 &  & LagNet & 0.436 & 0.414 & 0.416 & 0.408 & 0.409 \\
\addlinespace[0.2em]
 & 720 & Linear & 0.512 & 0.495 & 0.488 & 0.487 & 0.487 \\
 &  & LagLayer & 0.498 & 0.496 & 0.495 & 0.494 & 0.496 \\
 &  & LagNet & 0.494 & 0.493 & 0.469 & 0.480 & 0.469 \\
\midrule
ETTm2 & 96 & Linear & 0.202 & 0.190 & 0.183 & 0.182 & 0.181 \\
 &  & LagLayer & 0.195 & 0.192 & 0.184 & 0.186 & 0.182 \\
 &  & LagNet & 0.199 & 0.192 & 0.184 & 0.184 & 0.181 \\
\addlinespace[0.2em]
 & 192 & Linear & 0.265 & 0.253 & 0.248 & 0.246 & 0.247 \\
 &  & LagLayer & 0.259 & 0.254 & 0.250 & 0.248 & 0.248 \\
 &  & LagNet & 0.254 & 0.249 & 0.249 & 0.246 & 0.243 \\
\addlinespace[0.2em]
 & 336 & Linear & 0.324 & 0.315 & 0.309 & 0.308 & 0.308 \\
 &  & LagLayer & 0.316 & 0.316 & 0.313 & 0.316 & 0.314 \\
 &  & LagNet & 0.314 & 0.314 & 0.308 & 0.309 & 0.302 \\
\addlinespace[0.2em]
 & 720 & Linear & 0.422 & 0.416 & 0.410 & 0.410 & 0.408 \\
 &  & LagLayer & 0.417 & 0.413 & 0.412 & 0.413 & 0.411 \\
 &  & LagNet & 0.415 & 0.404 & 0.407 & 0.406 & 0.402 \\
\midrule
Electricity & 96 & Linear & 0.293 & 0.220 & 0.200 & 0.198 & 0.198 \\
 &  & LagLayer & 0.248 & 0.213 & 0.204 & 0.203 & 0.203 \\
 &  & LagNet & 0.217 & 0.188 & 0.176 & 0.171 & 0.167 \\
\addlinespace[0.2em]
 & 192 & Linear & 0.293 & 0.220 & 0.200 & 0.198 & 0.198 \\
 &  & LagLayer & 0.251 & 0.212 & 0.203 & 0.201 & 0.201 \\
 &  & LagNet & 0.209 & 0.191 & 0.181 & 0.177 & 0.174 \\
\addlinespace[0.2em]
 & 336 & Linear & 0.306 & 0.235 & 0.215 & 0.212 & 0.212 \\
 &  & LagLayer & 0.266 & 0.226 & 0.217 & 0.215 & 0.215 \\
 &  & LagNet & 0.223 & 0.208 & 0.197 & 0.192 & 0.189 \\
\addlinespace[0.2em]
 & 720 & Linear & 0.345 & 0.276 & 0.256 & 0.254 & 0.254 \\
 &  & LagLayer & 0.307 & 0.269 & 0.259 & 0.257 & 0.257 \\
 &  & LagNet & 0.293 & 0.250 & 0.244 & 0.234 & 0.233 \\
\midrule
Traffic & 96 & Linear & 0.949 & 0.720 & 0.655 & 0.647 & 0.645 \\
 &  & LagLayer & 0.801 & 0.702 & 0.666 & 0.656 & 0.653 \\
 &  & LagNet & 0.633 & 0.502 & 0.461 & 0.451 & 0.442 \\
\addlinespace[0.2em]
 & 192 & Linear & 0.940 & 0.675 & 0.607 & 0.600 & 0.599 \\
 &  & LagLayer & 0.776 & 0.650 & 0.619 & 0.608 & 0.607 \\
 &  & LagNet & 0.581 & 0.491 & 0.462 & 0.450 & 0.440 \\
\addlinespace[0.2em]
 & 336 & Linear & 0.936 & 0.680 & 0.616 & 0.607 & 0.606 \\
 &  & LagLayer & 0.777 & 0.661 & 0.632 & 0.618 & 0.616 \\
 &  & LagNet & 0.573 & 0.498 & 0.478 & 0.459 & 0.457 \\
\addlinespace[0.2em]
 & 720 & Linear & 0.968 & 0.728 & 0.653 & 0.645 & 0.644 \\
 &  & LagLayer & 0.805 & 0.708 & 0.671 & 0.656 & 0.654 \\
 &  & LagNet & 0.639 & 0.531 & 0.507 & 0.501 & 0.489 \\
\bottomrule
\end{tabular}
\end{tabular}
}
\caption{Complete horizon-wise sample-efficiency MSE results. Training fractions are applied only to the training windows, while validation and test splits remain unchanged.}
\label{tab:supp-sample-efficiency-mse}
\end{table*}

\section{Extended Structural Diagnostics}
\label{app:diagnostics}

\subsection{Metrics and Visualization Protocol}

This paper uses diagnostics to connect forecasting accuracy with learned operator structure.
Unless otherwise stated, the structural visualizations and adaptive-scaling diagnostics use the fixed diagnostic seed 2027.
For the cross-architecture visualization in the main paper, we define the mean history-to-future Jacobian $\bar{\boldsymbol{J}}\in\mathbb{R}^{H\times L}$ as
\begin{equation}
    \bar{j}_{h,\ell}
    =
    \frac{1}{N_JC}\sum_{n=1}^{N_J}\sum_{c=1}^{C}
    \frac{\partial \hat{y}_{n,c,h}}{\partial x_{n,c,\ell}},
\end{equation}
where $N_J$ sampled test windows and $C$ channels are averaged.
For nonlinear models, this input--output Jacobian captures local, input-dependent sensitivity rather than a fixed global forecasting matrix.
The sampled windows are evenly spaced over the test set, with
$N_J=\min(64,\lceil512/C\rceil)$, targeting approximately 512 channel traces
while capping the number of sampled windows at 64. Each mean Jacobian is divided by the 99.5th
percentile of its absolute entries, clipped to $[-1,1]$, and displayed with
a symmetric logarithmic color normalization. This panel-wise normalization
is used only to expose spatial structure; values are not compared across
panels.
For an induced forecasting matrix $\boldsymbol{W}\in\mathbb{R}^{H\times L}$ with scalar elements $w_{h,\ell}$, history-position importance is computed as
\begin{equation}
    I(\ell)
    =
    \frac{1}{H}\sum_{h=1}^{H}|w_{h,\ell}|.
\end{equation}
Forecast-step error is computed over all $N_{\mathrm{test}}$ test windows as
\begin{equation}
    \mathrm{MSE}(h)=
    \frac{1}{N_{\mathrm{test}}C}\sum_{n=1}^{N_{\mathrm{test}}}\sum_{c=1}^{C}
    \left(\hat{y}_{n,c,h}-y_{n,c,h}\right)^2.
\end{equation}
The main-paper adaptive-scaling diagnostic compares this forecast-step-resolved error with the learned horizon-wise coefficient $\alpha_h$ and separately examines the history-wise profile $\{\beta_\ell\}_{\ell=1}^{L}$.
Because $h$ indexes future steps whereas $\ell$ indexes historical positions, $\boldsymbol{\beta}$ is not interpreted through pointwise alignment with forecast-step MSE.
Instead, we examine whether its local response shape recurs qualitatively across repeated segments of the error curve, revealing nonuniform relative historical-position importance.
Related periodic error variation and structured scale responses were observed in several dataset--horizon settings; Traffic with $H=720$ is shown in the main paper as a representative case because these patterns are especially visible.
Parameter-matrix heatmaps are generated from learned parameters: dense operators are visualized directly, while lag-structured operators are reconstructed from lag kernels and optional scale coefficients.
Within each parameter-matrix comparison group, heatmaps use a shared color scale; if clipping is used for readability, the clipping percentile is reported in the figure caption.

\subsection{Complementary Magnitude-Only Visualizations}
\label{app:magnitude-counterparts}

\begin{figure}[!ht]
\centering
\includegraphics[width=0.78\columnwidth,draft=false]{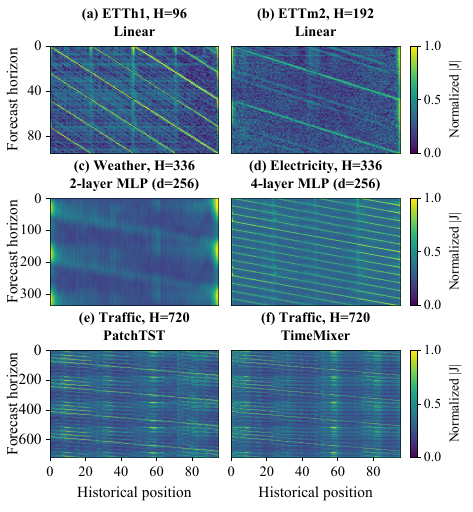}
\caption{Complementary magnitude-only history-to-future sensitivity visualization for the main-paper settings. Each panel averages absolute input--output Jacobians over test windows and channels.}
\label{fig:supp-main-jacobian-absolute}
\end{figure}

\begin{figure*}[p]
\centering
\includegraphics[width=0.98\textwidth,draft=false]{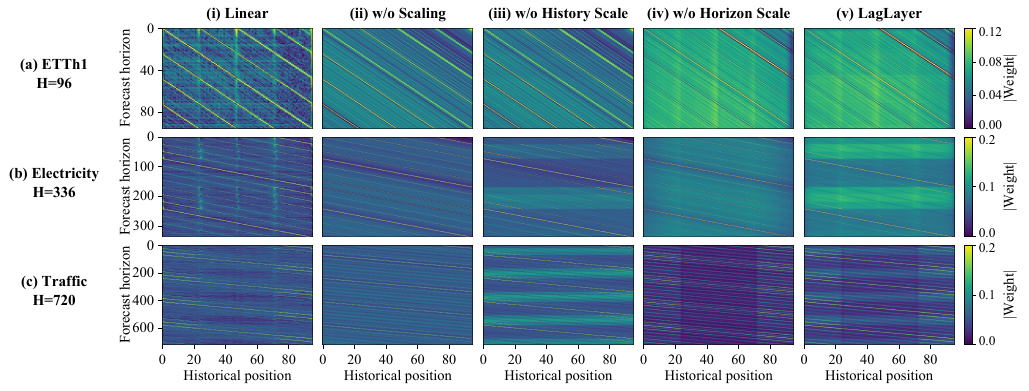}
\caption{Complementary magnitude-only forecasting-matrix visualization for the main-paper settings. The panels show absolute effective forecasting weights, and heatmaps within each row share a color scale clipped at its 99.5th percentile.}
\label{fig:supp-main-matrix-absolute}
\end{figure*}

For completeness, Figures~\ref{fig:supp-main-jacobian-absolute}
and~\ref{fig:supp-main-matrix-absolute} provide complementary magnitude-only
visualizations for the same settings as the two signed visualizations in the
main paper. These views discard coefficient signs and emphasize the strength
of the corresponding history-to-future connections.

\subsection{Additional Dense-Operator Motivation Visualizations}
\label{app:dense-motivation}

\begin{figure*}[p]
\centering
\includegraphics[width=0.965\textwidth,draft=false]{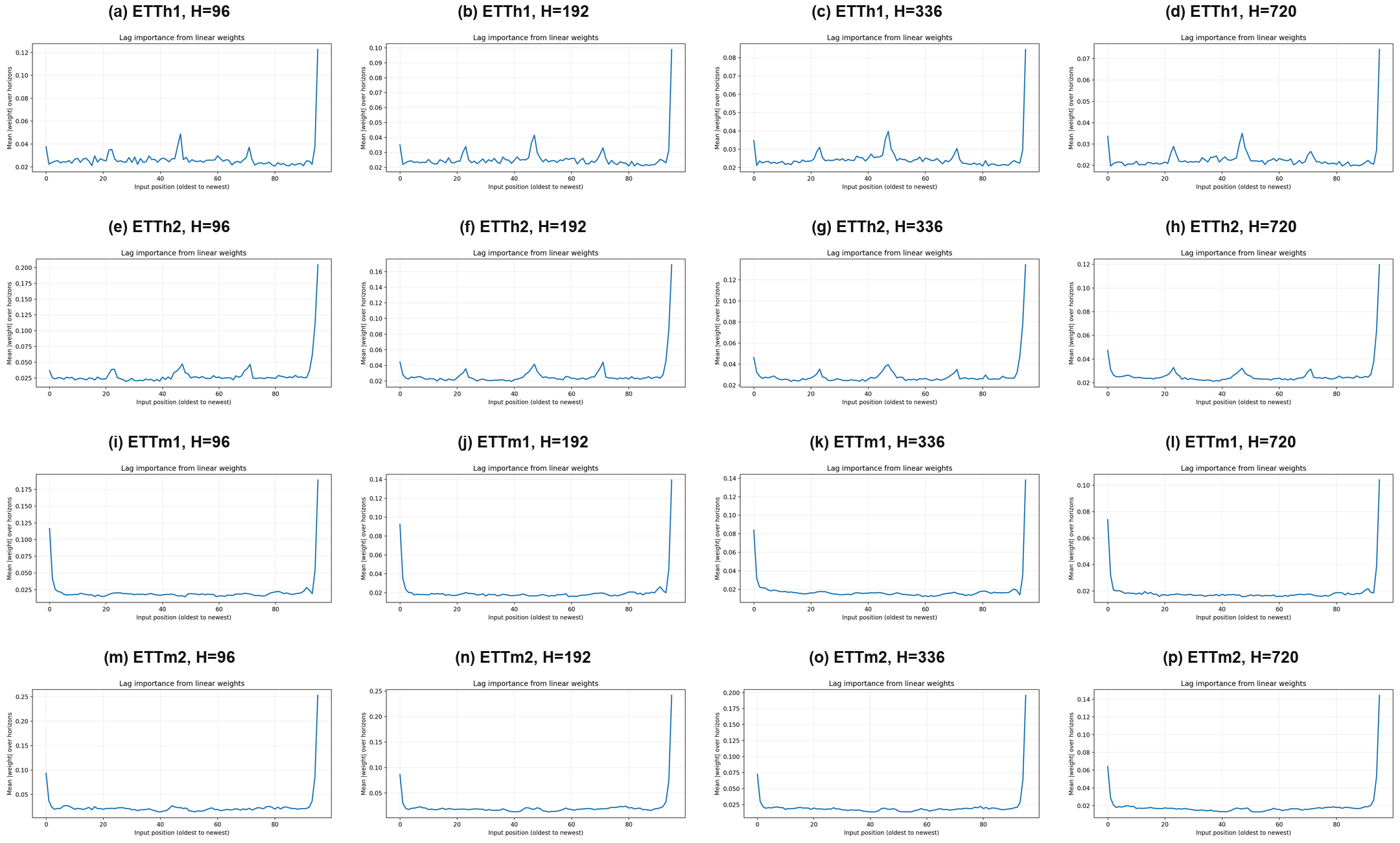}
\par
\includegraphics[width=0.965\textwidth,draft=false]{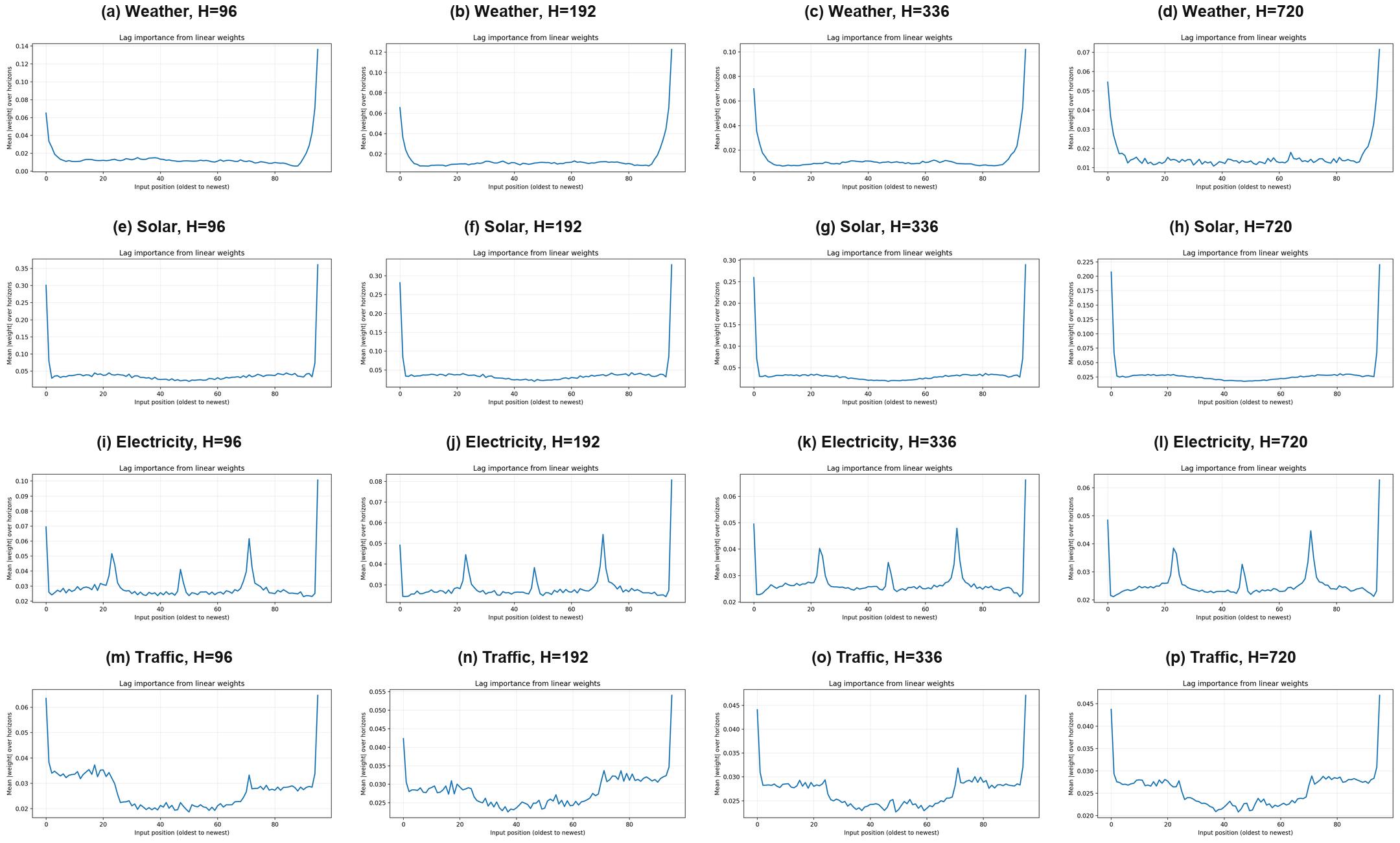}
\caption{History-position importance curves for Linear across all eight datasets and four prediction lengths. The upper panel contains the ETT datasets, and the lower panel contains Weather, Solar, Electricity, and Traffic. Each curve is computed as $I(\ell)=\frac{1}{H}\sum_{h=1}^{H}|w_{h,\ell}|$, without additional normalization. The nonuniform profiles show that historical positions can contribute with different overall strength.}
\label{fig:supp-dense-lag-all}
\end{figure*}

The main paper uses representative signed history-to-future sensitivity matrices across diverse datasets and forecasting architectures to motivate the lag-structured forecasting operator.
To provide broader evidence at the parameter level, the Linear columns of Figures~\ref{fig:supp-result-matrix-h96}--\ref{fig:supp-result-matrix-h720} show the complete absolute dense forecasting matrices on all eight benchmark datasets and all four prediction lengths $\{96,192,336,720\}$.
All models are trained with RevIN normalization, matching the model configuration used for the main-paper visualization.
Figure~\ref{fig:supp-dense-lag-all} summarizes the corresponding history-position importance curves, computed as
\[
    I(\ell)=\frac{1}{H}\sum_{h=1}^{H}|w_{h,\ell}|,
\]
without normalization, so the vertical axis reflects the actual average weight magnitude within each dataset.

Across datasets, these visualizations reveal diverse yet repeatedly structured
and nonuniform allocations across the history-to-future matrix. Together, they
support using relative lag as an explicit structural prior, while adaptive
scaling accommodates dataset- and position-dependent variation in connection
strength.

\subsection{Additional Forecasting Matrix Comparisons}
\label{app:matrix-comparisons}

\begin{figure*}[t]
\centering
\includegraphics[width=0.98\textwidth,draft=false]{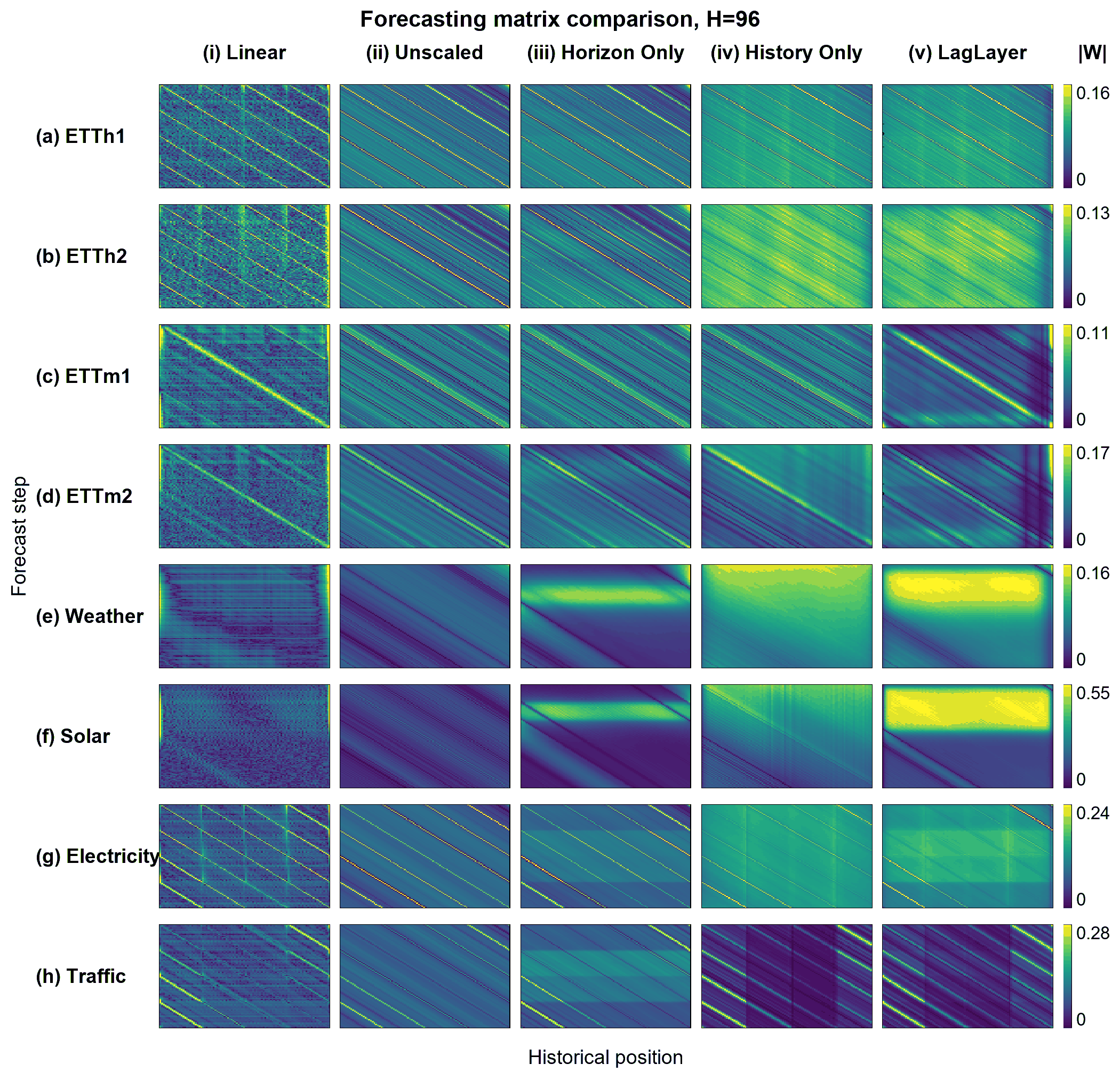}
\caption{Complete forecasting-matrix comparison for prediction length $H=96$. Panels show absolute effective forecasting weights. Rows correspond to datasets and columns correspond to operator variants. The horizontal and vertical axes span historical positions 1--96 and forecast steps 1--96, respectively. Each row uses a shared color scale across the five variants.}
\label{fig:supp-result-matrix-h96}
\end{figure*}

\begin{figure*}[t]
\centering
\includegraphics[width=0.98\textwidth,draft=false]{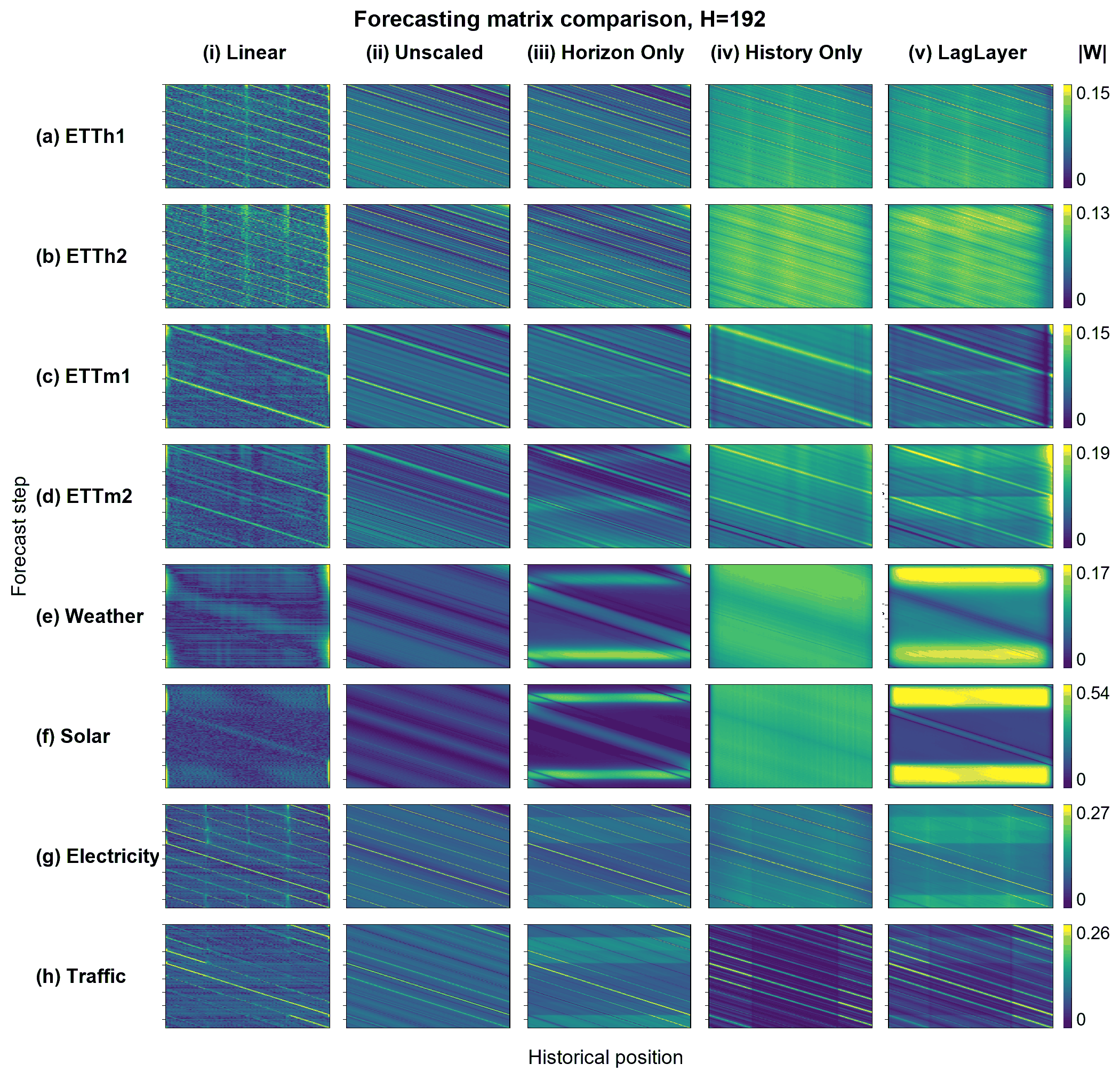}
\caption{Complete forecasting-matrix comparison for prediction length $H=192$. Panels show absolute effective forecasting weights. Rows correspond to datasets and columns correspond to operator variants. The horizontal and vertical axes span historical positions 1--96 and forecast steps 1--192, respectively. Each row uses a shared color scale across the five variants.}
\label{fig:supp-result-matrix-h192}
\end{figure*}

\begin{figure*}[t]
\centering
\includegraphics[width=0.98\textwidth,draft=false]{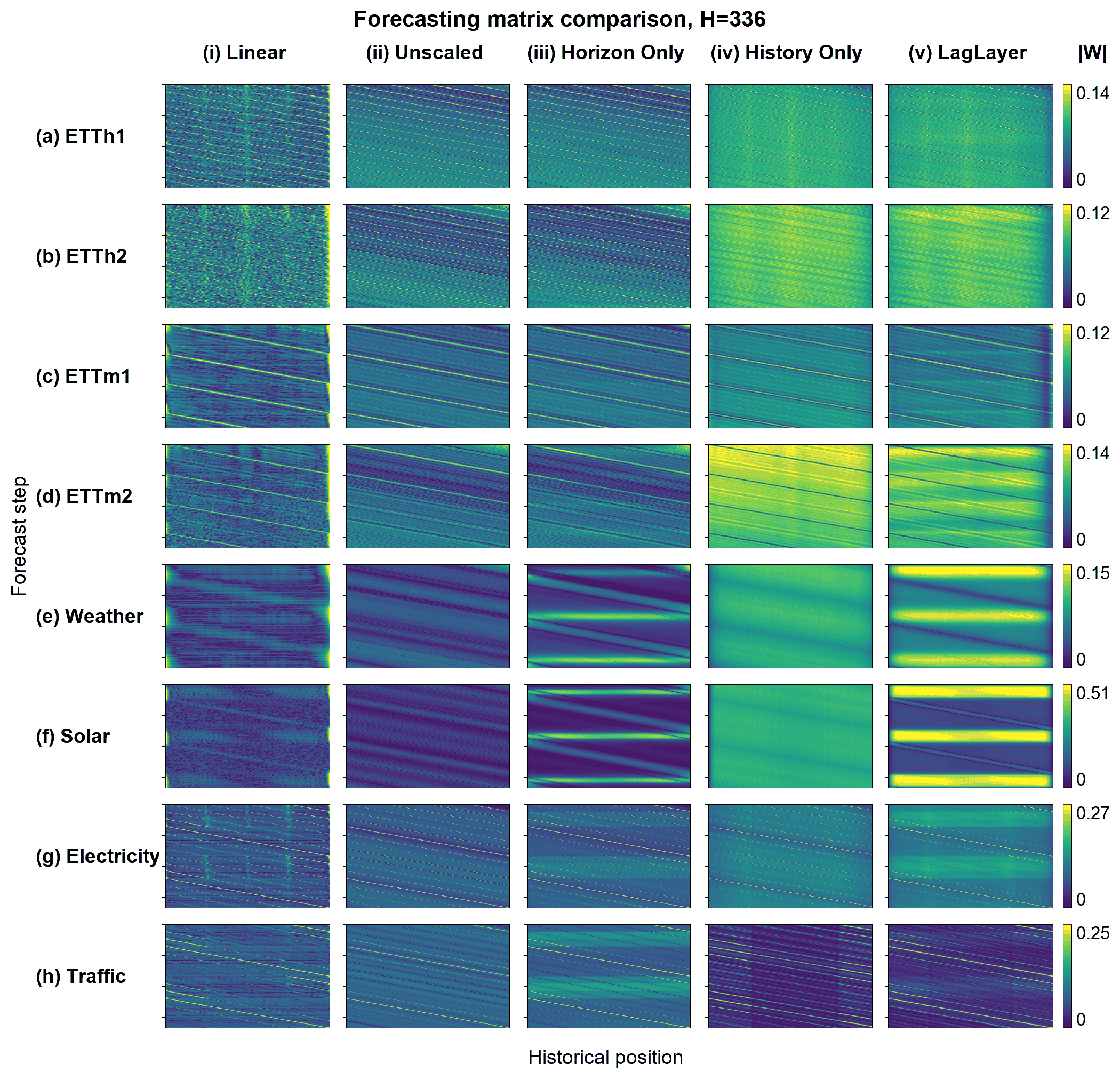}
\caption{Complete forecasting-matrix comparison for prediction length $H=336$. Panels show absolute effective forecasting weights. Rows correspond to datasets and columns correspond to operator variants. The horizontal and vertical axes span historical positions 1--96 and forecast steps 1--336, respectively. Each row uses a shared color scale across the five variants.}
\label{fig:supp-result-matrix-h336}
\end{figure*}

\begin{figure*}[t]
\centering
\includegraphics[width=0.98\textwidth,draft=false]{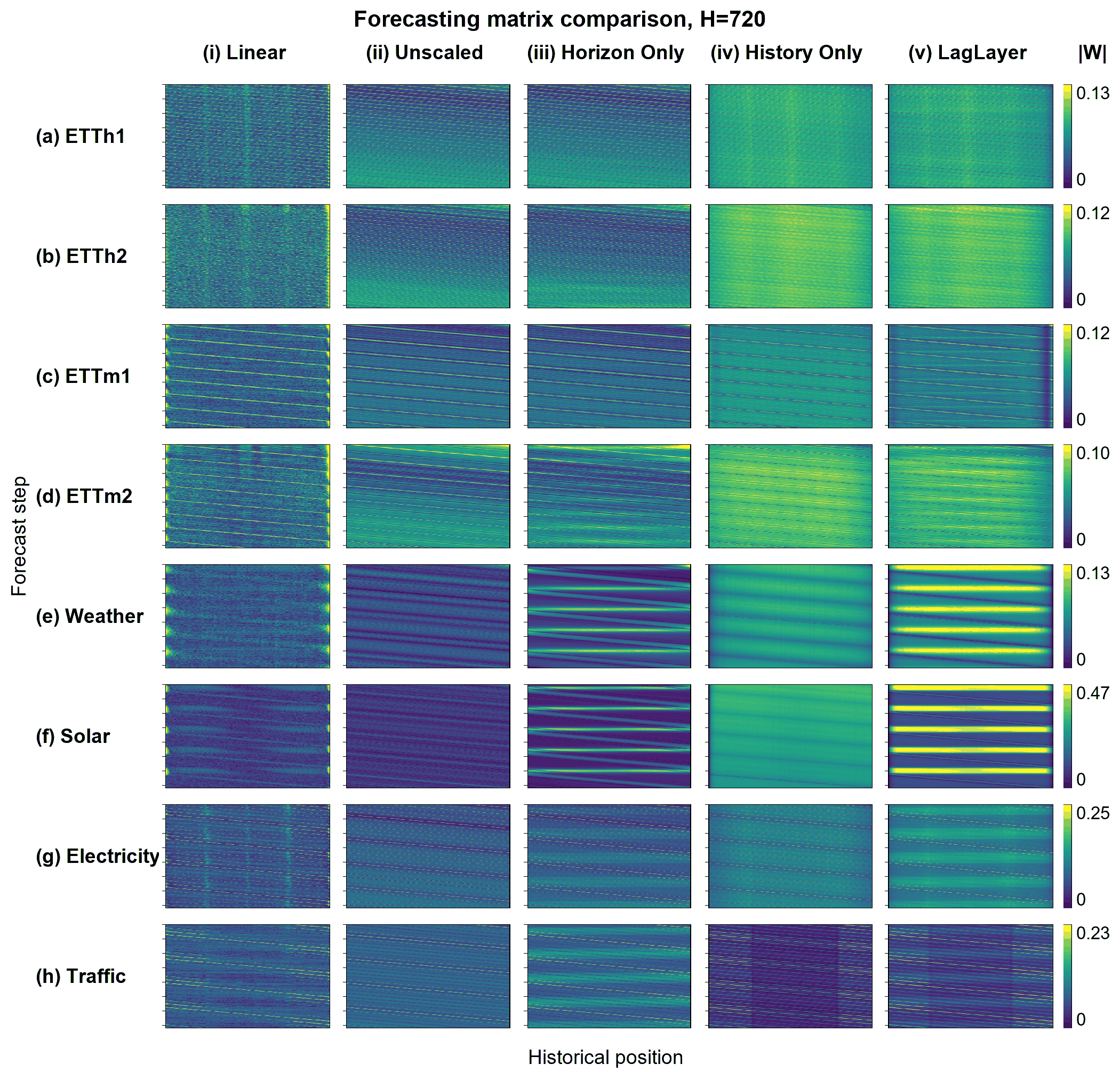}
\caption{Complete forecasting-matrix comparison for prediction length $H=720$. Panels show absolute effective forecasting weights. Rows correspond to datasets and columns correspond to operator variants. The horizontal and vertical axes span historical positions 1--96 and forecast steps 1--720, respectively. Each row uses a shared color scale across the five variants.}
\label{fig:supp-result-matrix-h720}
\end{figure*}

The main paper visualizes representative signed forecasting matrices for three dataset--horizon pairs.
Figures~\ref{fig:supp-result-matrix-h96}--\ref{fig:supp-result-matrix-h720} provide the complete set of matrix comparisons for all eight benchmark datasets and all four prediction lengths.
Each figure fixes one prediction length and compares five operator variants: Linear, \laglayer{} without scaling, the two one-sided scale ablations, and the complete \laglayer{} with both scales.
Within each row, the five heatmaps use the same color scale, whose upper limit
is the 99.5th percentile of the absolute weights pooled across the five
variants. Weight magnitudes are therefore comparable within the same dataset
and horizon.
These complete supplementary panels intentionally show absolute effective
forecasting weights, complementing the representative signed comparison in
the main paper.
Across dataset--horizon settings, the scaled operators repeatedly exhibit horizon-wise bands and nonuniform history-wise modulation, providing broader matrix-level context for the representative coefficient profiles analyzed in the main paper.

\section{Limitations and Broader Impact}
\label{app:limitations}

\noindent\textbf{Limitations.}
\method{} is based on a relative-lag sharing prior.
This prior is natural for many sliding-window forecasting settings, but it may be less suitable when forecasting relations depend strongly on absolute calendar positions, regime changes, or non-repeating events.
Adaptive scaling alleviates part of this rigidity through row-wise and column-wise scale modulation, but it does not fully replace explicit calendar modeling or regime-aware modeling.
The present study focuses on temporal operator structure; integrating \laglayer{} with dedicated cross-variate interaction modules remains a direction for future work.

\noindent\textbf{Broader impact.}
This work studies a general forecasting architecture and does not target a specific high-risk application.
Potential benefits include more efficient and interpretable forecasting models, which may help deployment in resource-constrained environments.
As with other forecasting systems, inappropriate use can lead to poor decisions when the data distribution shifts, when uncertainty is ignored, or when forecasts are treated as deterministic guarantees.
Applications in safety-critical domains should therefore include uncertainty estimation, domain validation, and human oversight.

\bibliography{references}